\documentclass[times, review, 10pt]{elsarticle}

\usepackage[T1]{fontenc}    % use 8-bit T1 fonts
\usepackage{url}            % simple URL typesetting
\usepackage{booktabs}       % professional-quality tables
\usepackage{amsfonts}       % blackboard math symbols
\usepackage{nicefrac}       % compact symbols for 1/2, etc.
\usepackage{microtype}      % microtypography
\usepackage{xcolor}         % colors
\usepackage{graphicx}
\usepackage{wrapfig}
\usepackage{colortbl}
\usepackage{hyperref}       % hyperlinks

\newcommand{\etal}{\emph{et al. }} % notation of `w.r.t.`

\newcommand{\mod}[1]{{\color{black} #1}}

\journal{Pattern Recognition}

\begin{document}

\begin{frontmatter}

%% Title, authors and addresses

%% use the tnoteref command within \title for footnotes;
%% use the tnotetext command for theassociated footnote;
%% use the fnref command within \author or \affiliation for footnotes;
%% use the fntext command for theassociated footnote;
%% use the corref command within \author for corresponding author footnotes;
%% use the cortext command for theassociated footnote;
%% use the ead command for the email address,
%% and the form \ead[url] for the home page:
%% \title{Title\tnoteref{label1}}
%% \tnotetext[label1]{}
%% \author{Name\corref{cor1}\fnref{label2}}
%% \ead{email address}
%% \ead[url]{home page}
%% \fntext[label2]{}
%% \cortext[cor1]{}
%% \affiliation{organization={},
%%             addressline={},
%%             city={},
%%             postcode={},
%%             state={},
%%             country={}}
%% \fntext[label3]{}

\title{Revisiting the Generalization Problem of Low-level Vision Models Through the Lens of Image Deraining}

\author[affil1,affil2]{Jinfan Hu\fnref{fn1}}        
\ead{jf.hu1@siat.ac.cn}
\author[affil3,affil1]{Zhiyuan You\fnref{fn1}}
\ead{zhiyuanyou@foxmail.com}
\author[affil4]{Jinjin Gu}
\ead{jinjin.gu@insait.ai}
\author[affil5,affil6]{Kaiwen Zhu}
\ead{sqzhukaiwen@sjtu.edu.cn}
\author[affil3]{Tianfan Xue}
\ead{tfxue@ie.cuhk.edu.hk}
\author[affil7,affil8]{Chao Dong\corref{cor1}}
\ead{chao.dong@siat.ac.cn}

\fntext[fn1]{These authors contributed equally to this work, listed in alphabetical order.} 

\cortext[cor1]{Corresponding author.} 

\affiliation[affil1]{%
  organization={Shenzhen Institutes of Advanced Technology, Chinese Academy of Sciences},
  city={Shenzhen},
  postcode={518055},
  country={China}
}

\affiliation[affil2]{%
  organization={University of Chinese Academy of Sciences},
  city={Beijing},
  postcode={100049},
  country={China}
}

\affiliation[affil3]{%
  organization={The Chinese University of Hong Kong},
  city={Hong Kong},
  postcode={999077},
  country={China}
}

\affiliation[affil4]{%
  organization={INSAIT, Sofia University “St. Kliment Ohridski”},
  city={Sofia},
  postcode={1784}, 
  country={Bulgaria}
}

\affiliation[affil5]{%
  organization={Shanghai Jiao Tong University},
  city={Shanghai},
  postcode={200240},
  country={China}
}

\affiliation[affil6]{%
  organization={Shanghai Artificial Intelligence Laboratory},
  city={Shanghai},
  postcode={200232},
  country={China}
}
\affiliation[affil7]{%
  organization={Shenzhen Key Lab of Computer Vision and Pattern Recognition, Shenzhen Institutes of Advanced Technology, Chinese Academy of Sciences},
  city={Shenzhen},
  postcode={518055},
  country={China}
}
\affiliation[affil8]{%
  organization={Shenzhen University of Advanced Technology},
  city={Shenzhen},
  postcode={518055},
  country={China}
}
%% Abstract
\begin{abstract}
\mod{Generalization to unseen degradations remains a fundamental challenge for low-level vision  models. This paper aims to investigate the underlying mechanism of this failure, using image deraining as a primary case study due to its well-defined and decoupled structure. Through systematic experiments, we reveal that generalization issues are not primarily caused by limited network capacity, but rather by a ``shortcut learning'' phenomenon driven by the relative complexity between image content and degradation patterns. We find that when background content is excessively complex, networks preferentially overfit the simpler degradation characteristics to minimize training loss, thereby failing to learn the underlying image distribution. To address this, we propose two principled strategies: (1) balancing the complexity of training data (backgrounds vs. degradations) to redirect the network's focus toward content reconstruction, and (2) leveraging strong content priors from pre-trained generative models to physically constrain the network onto a high-quality image manifold. Extensive experiments on image deraining, denoising, and deblurring validate our theoretical insights. Our work provides an interpretability-driven perspective and a principled methodology for improving the robustness and generalization of low-level vision models.}

\end{abstract}

%% Keywords
\begin{keyword}
Low-level Vision\sep Generalization\sep Generative Models\sep Interpretability\sep Shortcut Learning.
\end{keyword}

\end{frontmatter}

\section{Introduction}

Generalization challenges remain a fundamental constraint in deep learning, particularly in low-level vision (LV) where models trained on synthetic data often fail to capture natural image complexity. LV generalization is not a simple extension of high-level vision research. It is difficult to quantify when tangled with image content. This entanglement complicates assessments of reconstruction fidelity and generalization; therefore, we focus on additive degradations as a starting point.

Image deraining is selected as our primary case study due to its formulated linear superimposition and spatially separable nature, which allows for specifically masked quantitative evaluations of rain removal and background reconstruction. This clear separation makes deraining an intuitive starting point to isolate the generalization problem from extraneous factors. While we begin with deraining for its decoupled structure, our study further extends to other representative tasks, including image denoising and deblurring, to validate the broader applicability of our findings.

We argue that generalization failure results from networks overfitting training patterns, a problem rooted in the unproven assumption that larger, more diverse datasets inevitably improve performance. However, this approach does not effectively address the issue. We contend that excessive background complexity encourages networks to find "shortcuts": they overfit simpler degradation patterns instead of learning the difficult task of content reconstruction. By separately measuring background and rain removal, we arrive at counter-intuitive conclusions.

\begin{figure*}[t]
    \centering
    \includegraphics[width=\linewidth]{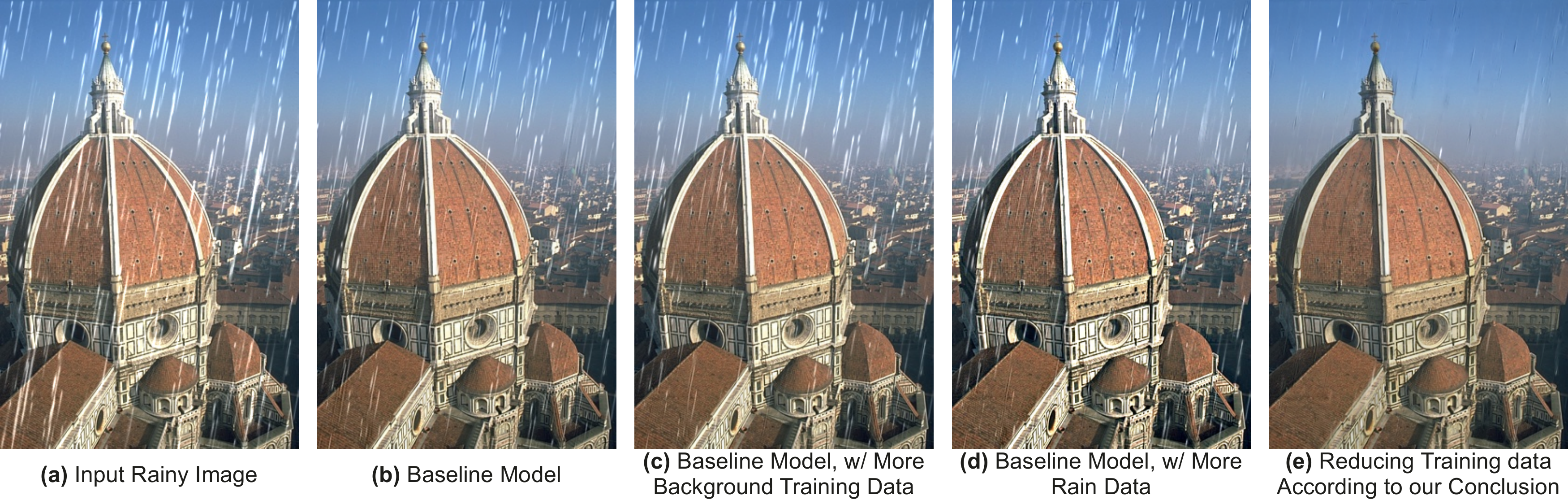}
   \vspace{-10mm}
    \caption{
    Existing models generalize poorly: after synthetic training, unseen rain \textbf{(a)} results in limited removal \textbf{(b)}. Conventional solutions like adding more background images \textbf{(c)} or rain patterns \textbf{(d)} fail to resolve this. We offer a counter-intuitive insight: using \emph{much less} background data \textbf{(e)} can enhance generalization.
    }
    \label{fig:teaser}
    \vspace{-3mm}
\end{figure*}

\textbf{Our key findings:} In separating content from degradation, deep networks preferentially model the less complex element as a shortcut to minimize training loss. Consequently, complex backgrounds with simple rain streaks lead to rain overfitting and failure on real-world data. Conversely, training on less complex backgrounds (\figurename~\ref{fig:teaser} (e)) makes the background the easier element to learn, incentivizing the network to focus on reconstruction. Crucially, our conclusion is not that less data is better, but that the relative complexity between content and degradation is the critical factor dictating learning and generalization. 
Improperly designed objectives create a "loophole" for networks to "slack off" by memorizing simple degradation instead of rich natural image patterns. This shortcut leads to poor performance on unseen data, establishing a core principle: \textbf{robust models should learn the image content distribution rather than specific degradation characteristics}.

This work significantly extends our NeurIPS 2023 paper \cite{gu2023networks} with several new contributions. First, we expand our analysis to show that background sharpness is crucial; networks learn low-sharpness content more readily, enhancing generalization. We supplement this with a vision-language model \cite{you2023depicting} for holistic evaluation. Second, a designed 1D "toy task" intuitively visualizes the network's general tendency to fit the simpler element in a mixture. Finally, we propose a strategy leveraging generative content priors to force networks to focus on content, validated through significant improvements across deraining, denoising, and deblurring tasks.

The rest of this paper is organized as follows: Section \ref{sec:related} reviews related work; Section \ref{sec:methodology} outlines our analysis framework; Section \ref{sec:understanding} presents key experimental insights; Section \ref{sec:analogous} introduces the toy task; Section \ref{sec:implict} proposes practical enhancement strategies; and Section \ref{sec:conclusion} concludes the paper.

\mod{\section{Related Works}\label{sec:related}

\begin{figure*}[t]
    \centering
    \includegraphics[width=\linewidth]{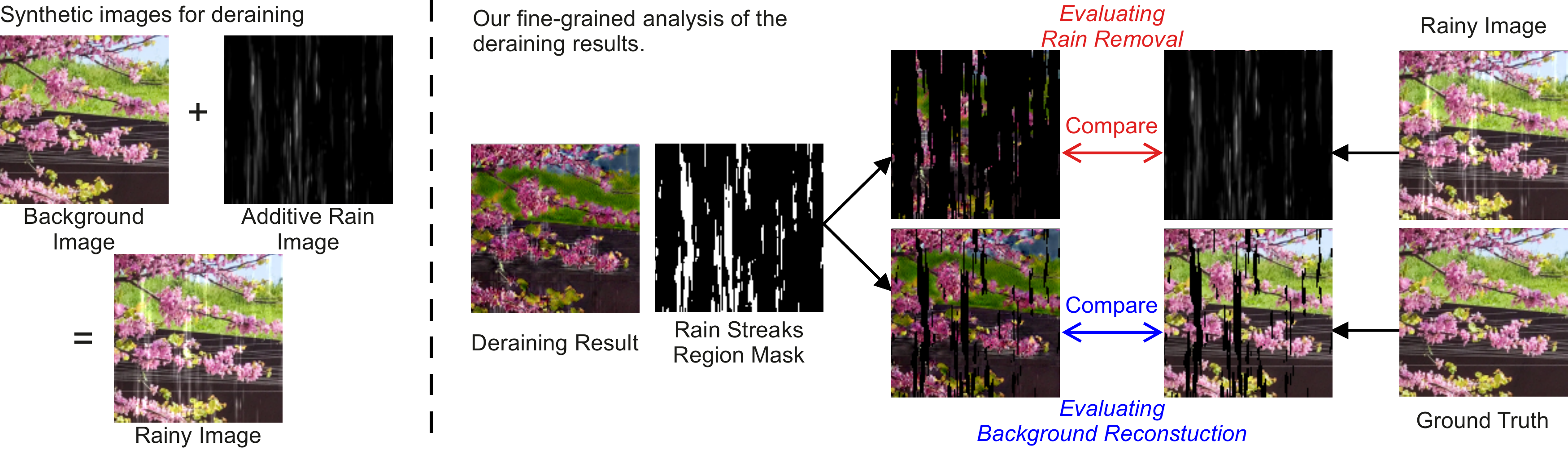}
    \vspace{-10mm}
    \caption{\textbf{(Left)} The illustration of the rainy image synthesis. \textbf{(Right)} Our pipeline of decoupled analysis.}
    \label{fig:output}
    \vspace{-3mm}
\end{figure*}

\textbf{Image Deraining and Generalization.}
Numerous methods have been proposed for state-of-the-art deraining \cite{zhang2023data,wen2026structure}. Early efforts focused on photo-realistic synthesis \cite{garg2006photorealistic} or manual editing \cite{zhang2018density,fu2017clearing}, yet models trained on synthetic data often fail in practice due to the domain gap. To mitigate this, researchers have contributed real-world datasets \cite{yang2017deep}, or semi-supervised approaches \cite{wei2019semi,yasarla2020syn2real}. While improving performance, these works largely bypass the core generalization problem by converting ``unseen'' to ``seen'' rain through data expansion. In contrast, we analyze the fundamental mechanism of generalization without proposing new datasets or architectures.

\textbf{Shortcut Learning and Interpretability.}
Our analysis is closely related to the concept of shortcut learning. Geirhos \etal \cite{geirhos2020shortcut} defined shortcut learning as the tendency of neural networks to learn shortcuts that perform well on benchmarks but fail on out-of-distribution data. While extensively studied in high-level vision, shortcuts remain under-explored in LV. Although interpretability tools exist for super-resolution \cite{gu2021interpreting,xie2021finding} and general restoration \cite{hu2024interpreting}, they primarily focus on inference or filter discrimination. To our knowledge, this work is the first to identify LV shortcuts through the lens of complexity competition between image content and degradation during training.

\textbf{Generative Priors and Guidance in Restoration.}
Generative priors from pre-trained foundations, such as VQGAN \cite{esser2021taming} and diffusion models \cite{podellsdxl}, have shown remarkable success in restoration \cite{liu2024adaptbir,zhou2022towards,lin2024diffbir} by leveraging high-quality (HQ) manifolds. Beyond pre-trained priors, researchers use explicit HQ information to guide restoration; for instance, Hqg-net \cite{he2023hqg} introduces HQ cues into medical image enhancement via variational normalization to ensure structural fidelity. Unlike these method-driven works that push SOTA performance, our study is analysis-driven. We utilize generative priors as probing tools rather than target algorithms. By freezing a pre-trained codebook, we effectively block degradation-learning shortcuts, empirically proving that prioritizing content priors is a fundamental solution to the generalization bottleneck.}

\section{Analysis of Image Deraining}\label{sec:methodology}
\subsection{Construction of Training Objective}
In this subsection, we analyze the generalization performance of different deraining models by setting a variety of training objectives in order to observe their effects.
The training data and the loss function jointly determine the training objective of a deep network.
As shown in \figurename~\ref{fig:output} (left), a rainy image $I$ can be modelled using a linear model $I=B+R$, where $B$ is the image background, and $R$ is the additive rain streaks.
We change the training objectives with different background images and rain streaks.

\subsubsection{Background Images.}
\begin{figure*}[t]
    \centering
    \includegraphics[width=\linewidth]{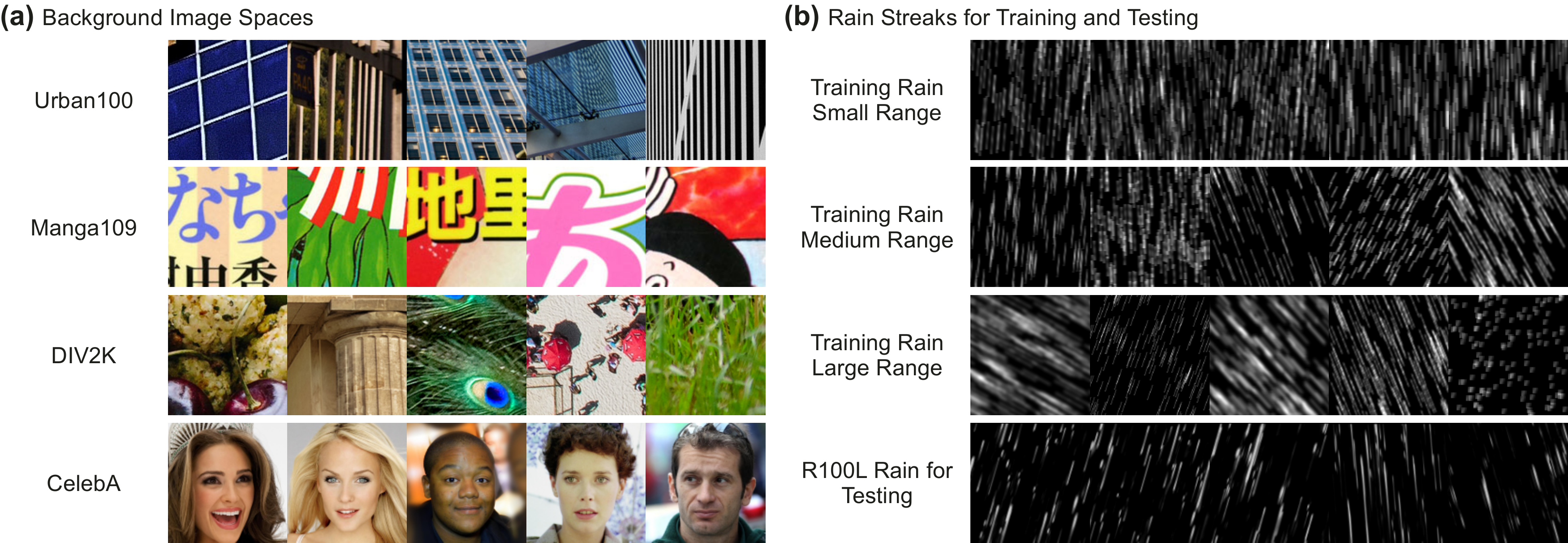}
    \vspace{-10mm}
    \caption{
\textbf{(a)} Background datasets with varying complexity: CelebA (faces), DIV2K (natural textures), Manga109 (sharp edges), and Urban100 (repetitive patterns). \textbf{(b)} Rain streaks used in our experiments.
    }
    \label{fig:data}
    \vspace{-3mm}
\end{figure*}

Conventional deraining methods typically utilize complex street views \cite{hsu2023recurrent} or natural images \cite{schaefer2003ucid,meng2025wavelet} based on the common belief that vast and complex data facilitates realistic reconstruction \cite{fu2017removing,zhang2018density}. We challenge this assumption by systematically varying training backgrounds in two ways. First, we vary the data scale, as it influences learning complexity. We hypothesize that excessive background complexity causes the network to overfit simpler degradation patterns instead of actual content. To observe behavior in extreme scenarios, we build training sets using 8, 16, 32, 64, 128, 256, 512, and 1024 patches ($128\times128$), along with a set of 30,000 patches for sufficient sampling. Second, we consider image content, as regular patterns are easier to fit than complex structures like faces \cite{bagrov2020multiscale}. We thus utilize four distinct sources to control content complexity: CelebA (faces) \cite{liu2015deep}, DIV2K (natural) \cite{timofte2017ntire}, Manga109 (comics) \cite{matsui2017sketch}, and Urban100 (buildings) \cite{huang2015single}, as shown in \figurename~\ref{fig:data} (a).

\begin{table}[t]
\caption{Different rain streaks settings.}\label{tab:rain}
\resizebox{0.7\linewidth}{!}{
\begin{tabular}{c|cccc}
\hline
    Range & Quantity & Width & Length & Direction \\
    \midrule
    Small & $[200, 300]$ & \{5\} & $[30, 31]$ & $[-5^{\circ},5^{\circ}]$ \\
    Medium & $[200, 300]$ & \{5,7,9\} & $[20, 40]$ & $[-30^{\circ}, 30^{\circ}]$ \\
    Large & $[200, 300]$ & \{1,3,5,7,9\} & $[5, 60]$ & $[-70^{\circ}, 70^{\circ}]$ \\
    \bottomrule
\end{tabular}}
\centering
\vspace{-3mm}
\end{table}

\subsubsection{Rain streaks synthesis.}
\label{sec:method:1:rain}
Since collecting real-world rainy/clean image pairs is difficult, we follow standard practice \cite{garg2006photorealistic,fu2017removing} and synthesize our data.
To properly evaluate generalization, we deliberately use different rain synthesis methods for training and testing.
For training, we use a computational model\footnote{The re-implementation of the PhotoShop rain streaks synthesis method. Please refer to \href{https://www.photoshopessentials.com/photo-effects/photoshop-weather-effects-rain/}{this link}.} to generate diverse rain streaks by varying their size, density, speed, and direction.
We create three distinct rain distributions for training to analyze their effect on generalization, as visualized in \figurename~\ref{fig:data} (b) and detailed in Tab. \ref{tab:rain}.
For testing, we use a different set of synthetic rain patterns from \cite{yang2017deep}.
Although the training and testing rain streaks look similar to humans, their underlying distributional shift presents a significant generalization challenge for the models.

\subsubsection{Loss Function.}
In low-level vision, the loss function is usually defined by the difference between the output image and the ground truth.
In our study, we use the $\ell_1$-norm loss, as it is the most commonly used and simplest loss function.

\subsection{Decoupling Analysis of Rain Removal Results}
\label{sec:method:2:fine}

\mod{Deraining models are typically evaluated using whole-image similarity metrics. However, these metrics can be misleading when assessing generalization to unseen degradations. For instance, a model that fails to generalize often outputs the input image directly. In this case, the background remains perfectly preserved, leading to a deceptively high PSNR, even though the rain streaks are not removed (see discussion in Sec. \ref{sec:prior}). To avoid this, we decouple the rainy input $I$ based on the additive model $I=B+R$. Since rain streaks $R$ are brighter than background $B$ (\figurename~\ref{fig:data} b), we generate a binary mask $M$ using a threshold $t$ ($M_{[i,j]}=1$ if $R_{[i,j]}>t$). This segments the output $\tilde{I}$ into rain regions ($\tilde{I}\odot M$) and background regions ($\tilde{I}\odot(1-M)$). We then define two decoupled metrics (Fig.~\ref{fig:output}). 

First, $E_R=\sqrt{\mathbb{E}[(\tilde{I}\odot M-I\odot M)^2]}$ denotes the \emph{Rain Removal Performance}. This metric measures the deviation of the output from the \textbf{rainy input} within streak regions. The rationale is that if a model fails to generalize, it preserves input streaks (resulting in $E_R \approx 0$); thus, a higher $E_R$ indicates successful suppression of unseen additive signals. 

Second, $E_B=\sqrt{\mathbb{E}[(\tilde{I}\odot (1-M)-B\odot (1-M))^2]}$ measures \emph{Background Reconstruction}. This metric compares non-rainy regions with the \textbf{ground truth} background $B$, serving as a fidelity check where smaller values indicate better preservation of image details.}

\subsection{Deep Models}
We categorize existing networks into three main groups. First, convolutional networks with deep residual connections, represented by ResNet \cite{ledig2017photo}. Second, encoder-decoder architectures like UNet \cite{ronneberger2015u}, which extract multi-scale features via down-sampling and up-sampling layers. Finally, we include Transformers characterized by self-attention, represented by SwinIR \cite{liang2021swinir}.

\section{Understanding Generalization}\label{sec:understanding}

\begin{figure*}[t]
    \centering
    \includegraphics[width=\linewidth]{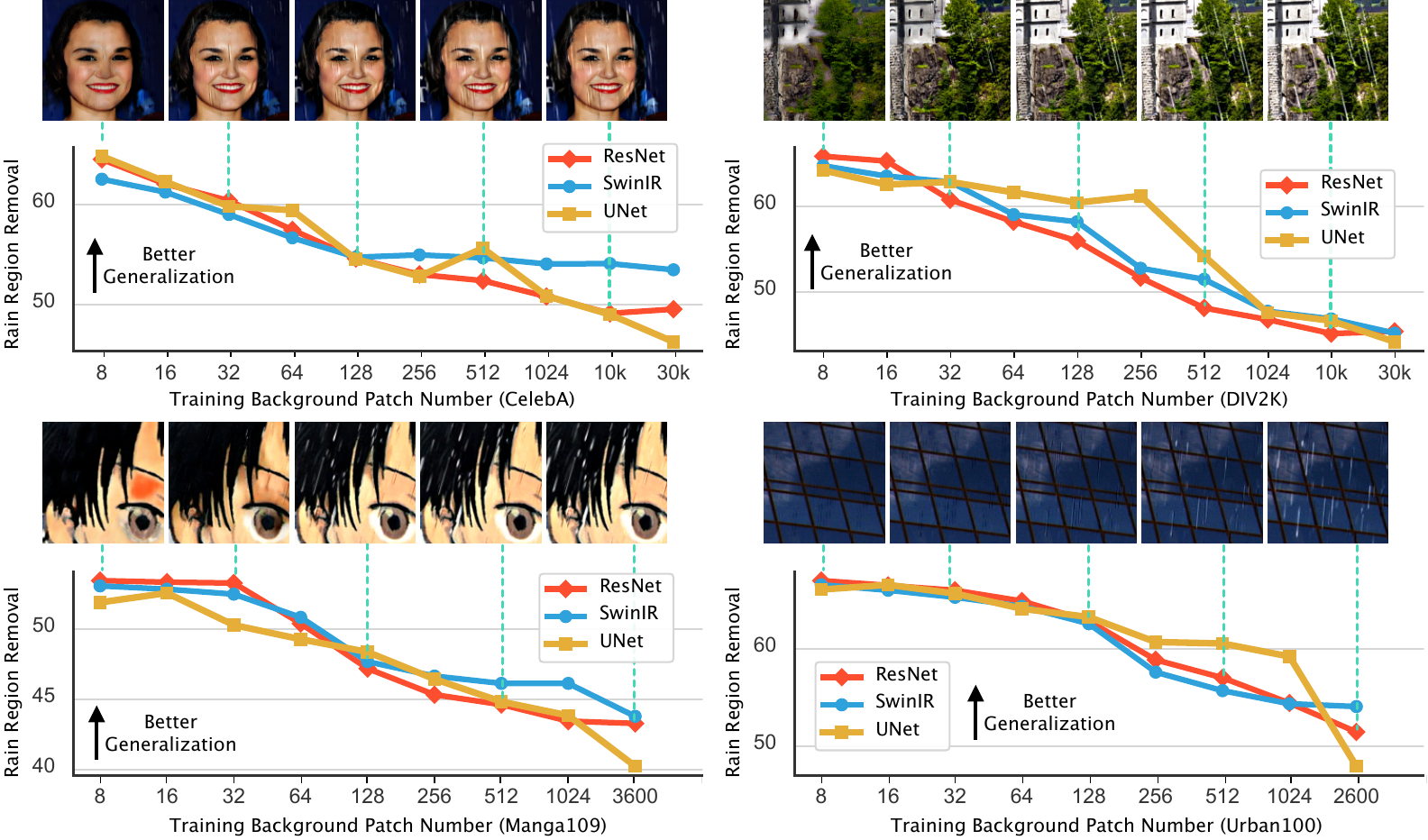}
    \vspace{-10mm}
    \caption{The relationship between the number of training patches and their rain removal performance. The $x$-axis represents the patch number, and the $y$-axis represents the rain removal effect $E_R$. Higher values on the $y$-axis mean better rain removal. The test rain patterns are not in the training set. The effect of rain removal at this time reflects the generalization performance. The qualitative results are obtained using ResNet.}
    \label{fig:rain-removal}
    \vspace{-3mm}
\end{figure*}

\subsection{Generalization on Rain Removal}
We conduct an analysis of the rain removal effect on unseen rain streaks.
Importantly, since we use different types of rain streak for training and testing, the results presented in this section all reflect generalization performance.
After extensive experimentation, we arrive at the following observations.

\subsubsection{Training with fewer background images leads to a better deraining effect.} \label{sec:fewer_image}
First, we fix the training rain to a medium level and vary only the background images to create different training objectives.
We conduct experiments on all four image categories, training separate models on sets with varying numbers of background patches.
We then evaluate these models on unseen test images, which use a different rain synthesis method \cite{yang2017deep} and backgrounds not seen during training.
The testing background images are sampled from the corresponding categories and are distinct from those in the training set.
The experimental results are presented in \figurename~\ref{fig:rain-removal}, revealing a consistent and counter-intuitive trend across all settings.
Remarkably, the deraining models trained on merely eight image patches can effectively handle unseen rain streaks.
Conversely, models trained on a large number of patches fail to remove the same rain streaks.
\emph{This observation deviates from conventional wisdom.}
More specifically, the model's rain removal ability deteriorates as the number of training background images increases, with performance collapsing almost entirely by 256 patches.
Beyond this point, adding more data (up to 30,000 patches) does not change the outcome: the models consistently fail to generalize.
Qualitative results visually confirm this trend.

Here, we attempt to elucidate this intriguing phenomenon.
To minimize the training loss, the network can adopt one of two strategies: learn to reconstruct the complex background content, or learn to remove the simpler degradation pattern.
Without explicit guidance, the network will always "opt" for the easier of these two tasks.
When trained on a large, diverse set of backgrounds, the content is far more complex than the repetitive rain streaks.
Consequently, the network takes a shortcut by learning to identify the rain, which leads to overfitting on the training degradation. This is why it fails when faced with unseen rain.
Conversely, when the background is simple (using only a few patches), learning the content becomes the easier task.
In this scenario, the network is incentivized to build a robust model of the background itself, without overfitting to the rain's features.
As a result, it generalizes effectively and removes unseen rain streaks.

\subsubsection{Training with less sharp backgrounds patches boosts deraining generalization.} \label{sec:sharpness}
We also propose that background sharpness is a key factor for generalization. 
Sharper images have more high-frequency content, encouraging the network to learn the simpler rain streaks as a shortcut.
To measure this, we quantify sharpness as the variance of the Laplacian operator's output on the grayscale image:
$ S = Var(Laplacian(I_{gray})) $.
Here, \( I_{gray} \) represents the grayscale image, and \(Var(\cdot)\) refers to the variance.
This operator highlights edges and textures, so a higher variance corresponds to a sharper image.

\begin{figure}[t]
    \centering
    \includegraphics[width=0.6\linewidth]{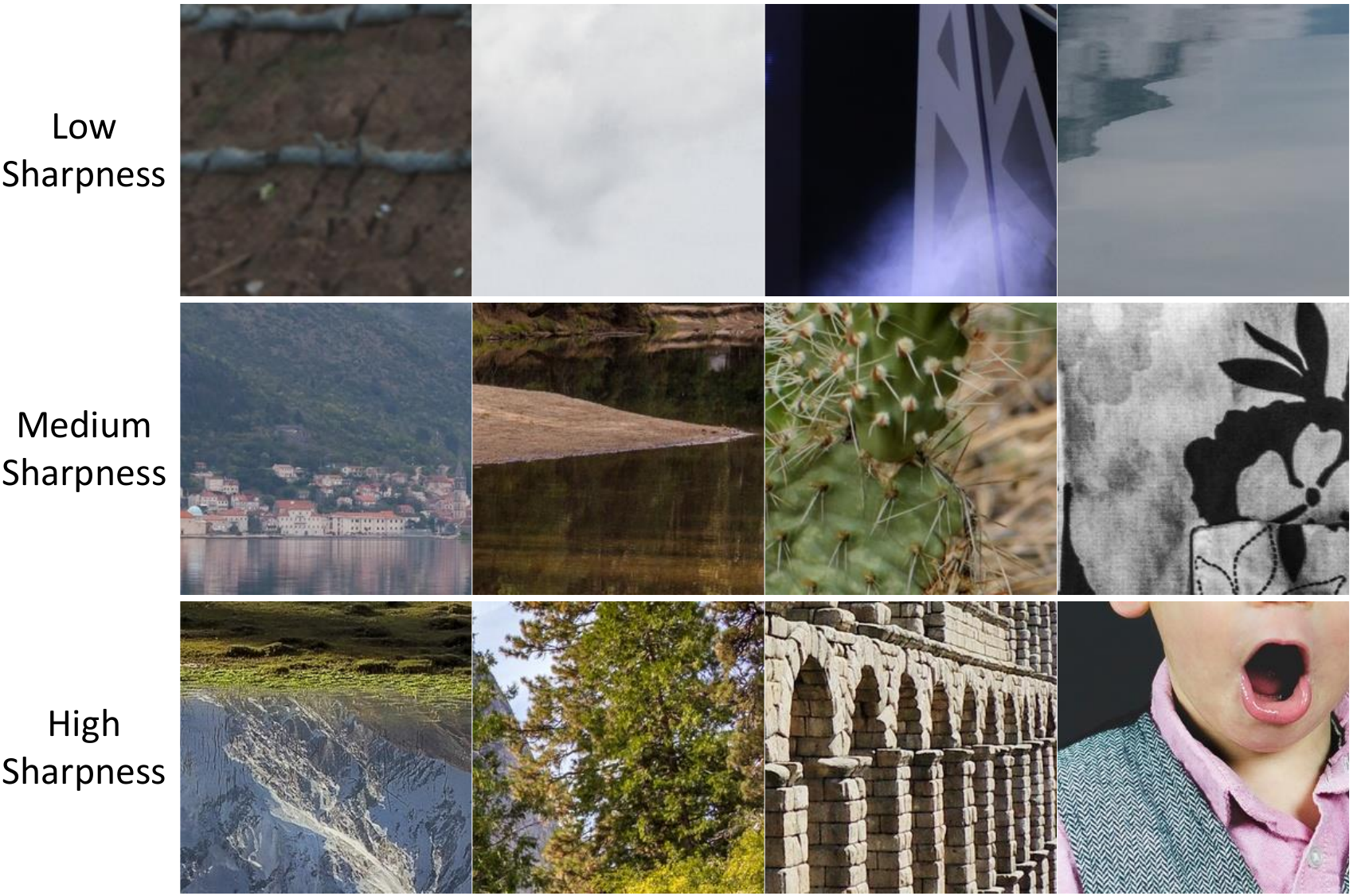}
    \vspace{-5mm}
    \caption{Examples from DIV2K classified as low, medium, and high sharpness.}
    \label{fig:sharpness}
\vspace{-3mm}
\end{figure}

\begin{figure}[t]
    \centering
    \includegraphics[width=0.6\linewidth]{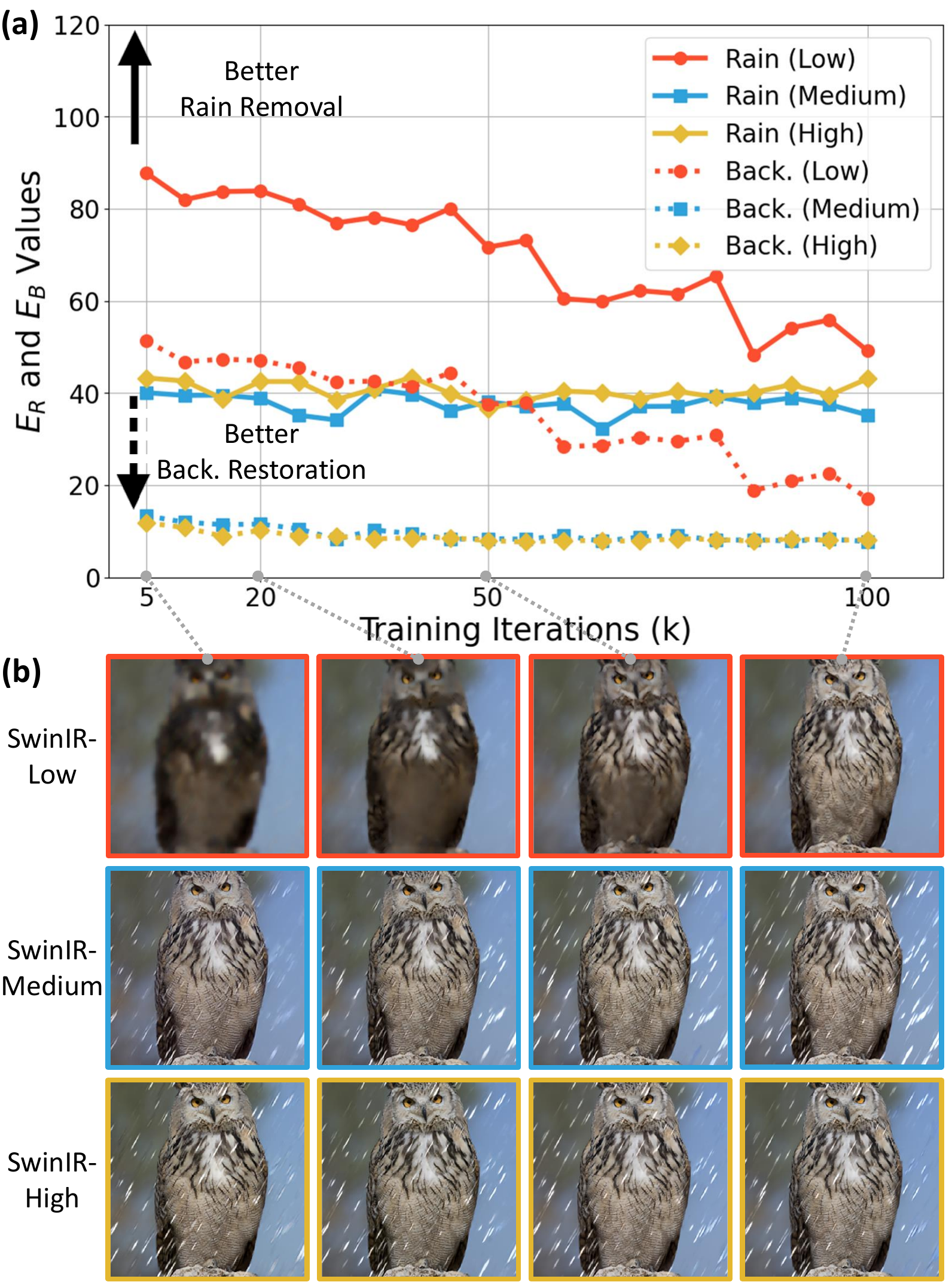}
    \vspace{-5mm}
    \caption{\textbf{(a)} The test performance of deraining generalization $E_R$ and background reconstruction $E_B$ of SwinIR, trained with training data of different sharpness levels. \textbf{(b)} Visual test results of models at different iterations. Zoom in for better comparison.}
    \label{fig:iter}
\vspace{-3mm}
\end{figure}

\begin{figure}[t]
    \centering
    \includegraphics[width=0.9\linewidth]{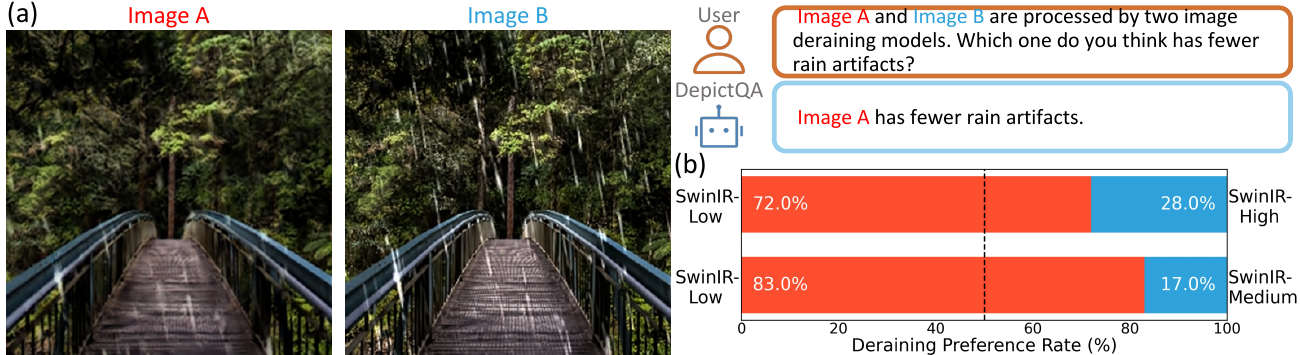}
\vspace{-5mm}
    \caption{\textbf{(a)} Framework for evaluating rain removal performance using DepictQA. 
    \textbf{(b)} Comparative results of different SwinIR models evaluated with DepictQA.}
    \label{fig:depictQA_sharpness}
\vspace{-3mm}
\end{figure}

A higher sharpness value indicates a detailed image, while a lower value suggests a blurrier or less distinct image.
We use the DIV2K dataset and group its images into three sharpness levels: low (< 50), medium (500 to 1000), and high (> 5000), with examples shown in Fig. \ref{fig:sharpness}.
We then create three distinct training sets, each containing 10,000 background images from one sharpness category.
For training, we use SwinIR for 100,000 iterations with the medium rain range.
All models are evaluated on a test set of 100 images featuring the Rain100L pattern, as seen in Fig. \ref{fig:data} (b).
The performance curves for rain removal and background reconstruction are presented in Fig. \ref{fig:iter} (a).

As illustrated in Fig. \ref{fig:iter} (a), the model trained on low-sharpness images consistently achieves the best generalization, outperforming the others in removing unseen rain at all stages of training.
However, this improved generalization comes at a cost: the network requires more iterations to learn the reconstruction of sharp backgrounds.

\begin{figure*}[t]
    \centering
    \includegraphics[width=\linewidth]{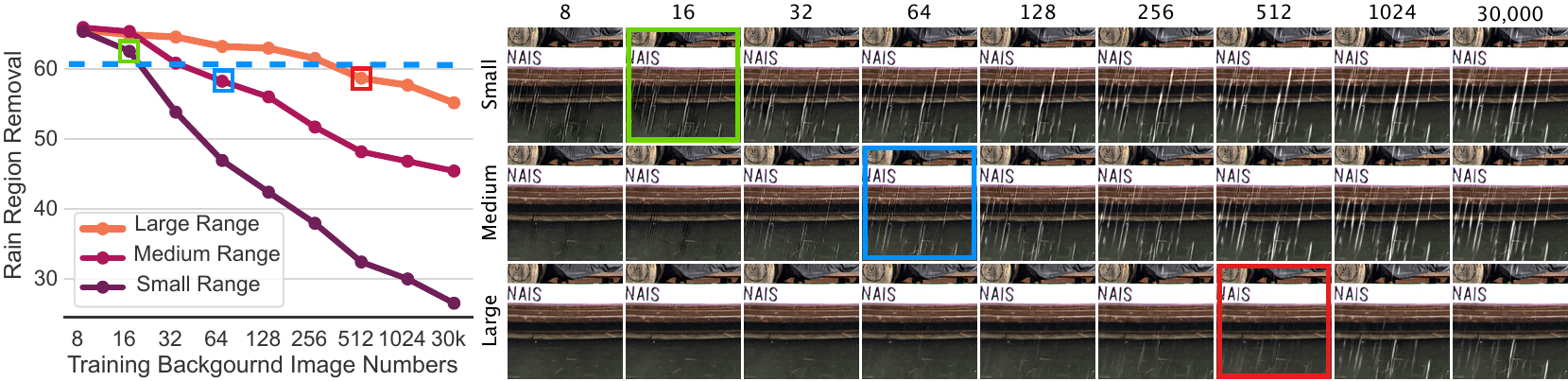}
\vspace{-10mm}
    \caption{When trained with different ranges, the model exhibits different effects. The $y$-axis represents the quantitative rain removal effect. When the rain removal performance is lowered to the blue dashed line, the qualitative effect of removing rain starts to decrease significantly. We use ResNet in this experiment.}
    \label{fig:rain-range}
\vspace{-3mm}
\end{figure*}

Second, the models' learning strategies diverge significantly as early as 5k iterations.
As shown in Fig. \ref{fig:iter} (b), SwinIR-Medium and SwinIR-High achieve sharper background reconstructions, but this comes at the cost of poor rain removal generalization.
It indicates that the networks fit the training rain streaks early on.
Additionally, their rain removal values \( E_R \) remain flat throughout the entire training process, as illustrated by the blue and yellow lines in Fig. \ref{fig:iter} (a).
In contrast, SwinIR-Low prioritizes learning the background content.
This results in less sharp reconstructions but superior generalization on the test set, as confirmed in the first row of Fig. \ref{fig:iter} (b).

Third, the performance of SwinIR-Low in Fig. \ref{fig:iter} (a) reveals a clear tradeoff between generalization and reconstruction. 
As training progresses, the network gets better at reconstructing the background, as shown in the first row of Fig. \ref{fig:iter} (b).
However, this improvement comes at the cost of generalization; the model begins to overfit the training rain, causing unseen rain streaks to become more apparent.
Concluding from Fig. \ref{fig:iter}, we can establish the network's fitting preference: low-sharpness content > training rain patterns > medium- and high-sharpness content.

In addition to quantitative metrics like $E_R$, we use the vision-language model DepictQA \cite{you2023depicting,you2024descriptive,zhu2024intelligent} for qualitative evaluation.
DepictQA effectively describes image quality linguistically, aligning with human perception.
We fine-tune DepictQA to focus solely on deraining performance, assessing how models generalize to unseen rain.
Fig. \ref{fig:depictQA_sharpness} (a) presents the image assessment paradigm.
We prompt DepictQA to compare two images and decide which contains fewer rain artifacts.
DepictQA then outputs its comparative judgment.
Fig. \ref{fig:depictQA_sharpness} (b) shows the comparison results for SwinIR-Low, SwinIR-Medium, and SwinIR-High across 100 test images.
These results confirm our conclusion that low-sharpness training data leads to better generalization.

\subsubsection{The background-rain complexity balance determines the network behavior.}
To validate our hypothesis, we now increase the complexity of the training rain, as described in Section \ref{sec:method:1:rain}.
Recall that with simpler, medium-range rain, the model's generalization failed when using just 64 background patches.
Our theory predicts that making the rain patterns more complex makes them a harder target for the network to overfit.
Consequently, the model should now maintain its generalization ability even when trained on a larger number of background patches.

The experimental results are depicted in \figurename~\ref{fig:rain-range}.
For all training rain types, generalization consistently worsens as the number of background patches increases.
With enough backgrounds (30,000 patches), all models fail to generalize, confirming they are overfitting to the training rain's distribution.
Crucially, the "tipping point" where generalization fails depends on the rain's complexity.
When trained with complex, large-range rain, the model's performance only drops after using more than 512 background patches.
Conversely, with simple, small-range rain, the model fails with as few as 16 patches.
These results strongly support our core claim: the network's behavior is dictated by the relative complexity between the background and the rain.
We can approximate these tipping points: large-range rain is more complex than 512 patches, while medium-range rain is more complex than 64.
In every case, the network simply takes a shortcut, opting to learn whichever component is relatively easier.

\begin{figure}[t]
    \centering
    \includegraphics[width=0.7\linewidth]{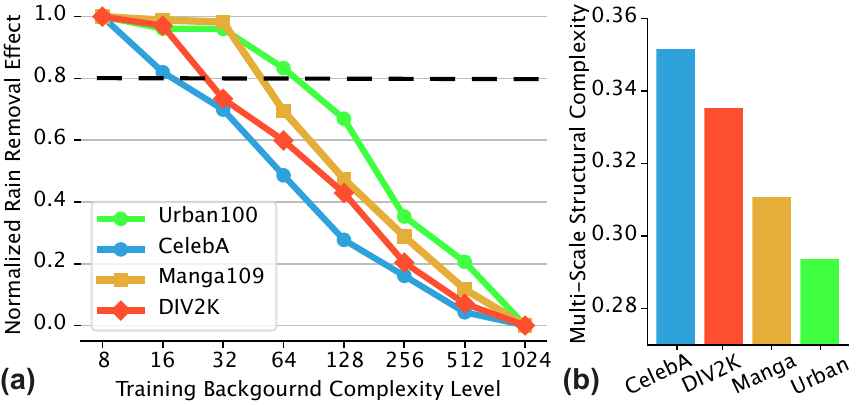} 
\vspace{-5mm}
    \caption{\mod{(a) The relationship between training patches amount and their normalized rain removal performance. When the value is lowered to the dashed line, the qualitative effect of removing rain starts to decrease significantly. (b) The averaged complexity of different image categories by \cite{bagrov2020multiscale}.}}
    \label{fig:complexity}
\vspace{-3mm}
\end{figure}

\mod{\subsubsection{Impact of Background Complexity on Learning Dynamics}
\label{sec:complexity_analysis}

Since defining data complexity from the perspective of a ``black-box'' network is inherently difficult, we utilize \textit{data diversity} and \textit{information density} as operational proxies (see Sec.~\ref{sec:fewer_image} and \ref{sec:sharpness}). AdditionBy fixing training rain levels and modifying background categories, we observe how these dimensions influence generalization. Normalized results in \figurename~\ref{fig:complexity} (a) reveal that identical patch counts across different categories yield distinct generalization trajectories.

For instance, CelebA exhibits a sharp performance decline when increasing from 8 to 16 patches, whereas DIV2K remains stable. Manga109 and Urban100 maintain generalization until exceeding 32 patches. These results suggest that higher background complexity prompts the network to reach its ``tipping point'' earlier, leading it to overfit simpler degradation patterns rather than reconstructing content. Consequently, the empirical learning complexity is ranked as: CelebA > DIV2K > Manga109 > Urban100.
}
\mod{This ordering mirrors human perceptual intuition: CelebA faces exhibit strong hierarchical structures; DIV2K features rich textures with simpler global layouts; Manga images contain intricate line art and text; and Urban images consist of repetitive, predictable patterns like grids. 

To corroborate this, we utilize the multi-scale structural complexity metric by Bagrov \etal \cite{bagrov2020multiscale}. This metric provides an objective benchmark for the structural richness of natural patterns. As shown in \figurename~\ref{fig:complexity} (b), the scores align perfectly with our proposed ranking, providing formal evidence that the rate of generalization decline is indeed governed by the inherent complexity of the background manifold.
}

\mod{
The empirical evidence presented in this section identifies a critical phenomenon we characterize as shortcut learning in low-level vision. Unlike shortcuts in high-level classification \cite{geirhos2020shortcut}, which often involve reliance on spurious contextual cues, the shortcut in image restoration stems from the network's inherent preference for modeling the simpler component within the additive mixture $I = B + R$. 
When the background complexity outweighs that of the degradation (\figurename~\ref{fig:complexity}), the network undergoes bias amplification: as the background manifold becomes more intricate, the generalization performance collapses at an accelerating rate. This suggests that the network's inherent preference for simpler elements is magnified by the increasing difficulty of the reconstruction task. Consequently, to minimize training loss efficiently, the model disproportionately prioritizes the low-complexity rain streaks over the complex image content. This confirms that the primary bottleneck for generalization is not a deficiency in data volume, but a strategic learning failure where the network preferentially fits the simpler manifold.

In summary, this section identifies complexity competition as the core driver of generalization failure in LV. Complexity, defined via data diversity and information density, aligns with mathematical structural entropy \cite{bagrov2020multiscale} and dictates fitting preferences. Unlike high-level vision shortcuts \cite{geirhos2020shortcut}, LV shortcut learning manifests as the network ``slacking off'' on complex content reconstruction to over-fit the simpler degradation manifold. This process is driven by bias amplification, where increasing background complexity triggers a collapse in generalization (Fig.~\ref{fig:rain-range}).}

\subsection{Reconstruction on Background}
\label{sec:back_recon}
\begin{figure*}[t]
    \centering
    \includegraphics[width=\linewidth]{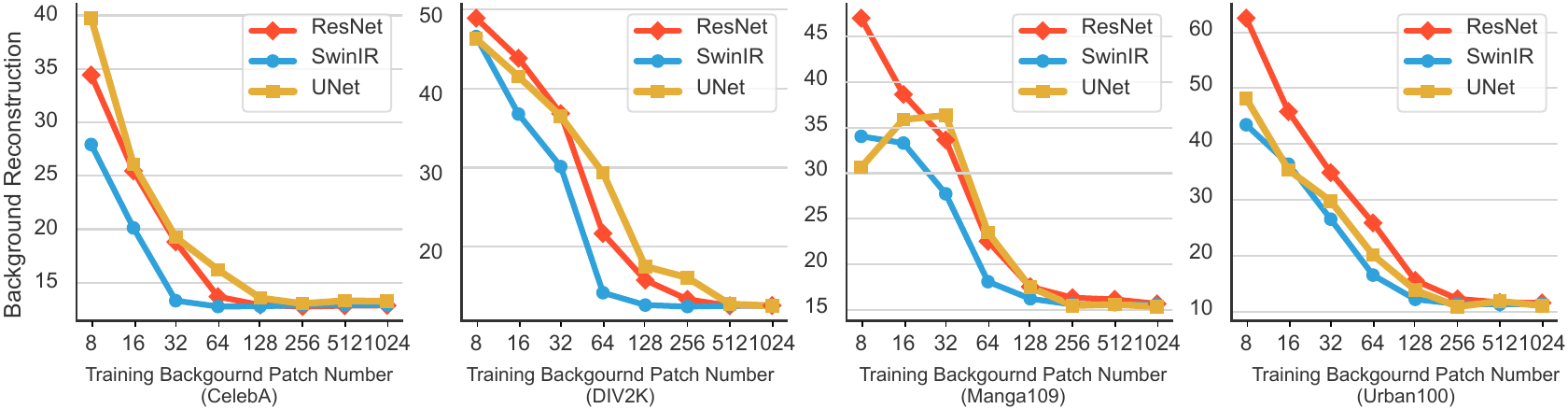}
    \vspace{-10mm}
    \caption{The relationship between the number of training patches and background reconstruction. For each plot, the $x$-axis represents the number, and the $y$-axis represents the background reconstruction error.}
    \label{fig:background_curve}
\vspace{-3mm}
\end{figure*}

The aforementioned results indicate that deraining capability can be enhanced by limiting the complexity of the training backgrounds.
However, we must emphasize that \textbf{\textit{our conclusion is not an argument for using less training data indiscriminately}}.
The content of the training data still critically impacts model performance, and in this section, we explore this importance.
Utilizing only a restricted set of backgrounds, while preventing overfitting to rain, introduces its own risk.
The network may conversely overfit to the limited selection of background images.
We now conduct investigations to address this particular concern.
Using the decoupled evaluation metric $E_B$ outlined in Section~\ref{sec:method:2:fine},  we can independently assess background reconstruction.
The results in \figurename~\ref{fig:background_curve} show that reconstruction quality improves as the number of training images increases.
More training data and sharper backgrounds do result in better image reconstruction, as observed in \figurename~\ref{fig:rain-removal} and \figurename~\ref{fig:iter} (b).
This highlights a critical tradeoff: more diverse data can enhance a network's foundational ability, but generalization will be compromised if the complexity between the background and the degradation is imbalanced.
Therefore, we argue that managing this balance, not just maximizing data quantity, is the key to improving performance and generalization.

\section{Further Validation with Analogous Task}\label{sec:analogous}
In this section, we further validate the network's "slacking off" behavior through a more intuitive experiment.
We designed an analogous task that provides a simpler, more controllable scenario to clearly observe the network's behavior.

\subsection{Construction of Function Denoising}
Image deraining is fundamentally a task of recovering a ground truth (GT) signal from an observation with additive degradation.
Inspired by this, we create an analogous toy task with the same problem formulation.
Specifically, we use a univariate cosine function as the GT signal and additive Gaussian noise as the degradation.
The cosine function is chosen because its periodic nature makes reconstruction fidelity easy to analyze.
Similarly, Gaussian noise provides a simple and controllable degradation pattern.
The GT function is defined as follows:
\begin{equation}
y = f(x) = 10\cos\left(\frac{O\pi}{10}x\right), \quad x \in [-10, 10],
\end{equation}
where the order \( O \) controls the oscillation frequency of the function.
By varying \( O \), we can directly adjust the signal's complexity for our experiments.
The data is then corrupted by Gaussian noise \( n \) drawn from the distribution \( \mathcal{N}(\mu, \sigma) \).
The network is trained to recover the clean signal $(x, y)$ from the noisy input $(x, y + n)$.

\begin{table}[t]
\caption{Function denoising vs. image restoration: $(x, y + n)$ represents the noisy points, where $(x, y)$ denotes the true data, $n$ is the noise. Similarly, $B + R$ represents a low-quality image, with $B$ as the background and $R$ as the additive degradation.
}\label{tab:toy_example}

\resizebox{0.7\linewidth}{!}{
\begin{tabular}{c|cc}
\hline
    Tasks & Function Denoising & Image Restoration \\
    \midrule
    Background Content & Cosine Functions & Natural Images \\
    Degradation Type & Gaussian  Noise $ n \sim \mathcal{N} (\mu, \sigma)$ & Additive Degradation $R$ \\
    Training Sample & Function Segments & Image Patches  \\
    Inference Input & Noisy Data $(x,y+n)$ & Low-Quality Image $B+R$\\
    Inference Output & Denoised Data $(x,y)$ & Restored Image $B$ \\
    \bottomrule
\end{tabular}}
\centering
\end{table}

\begin{figure}[t]
    \centering
    \includegraphics[width=0.7\linewidth]{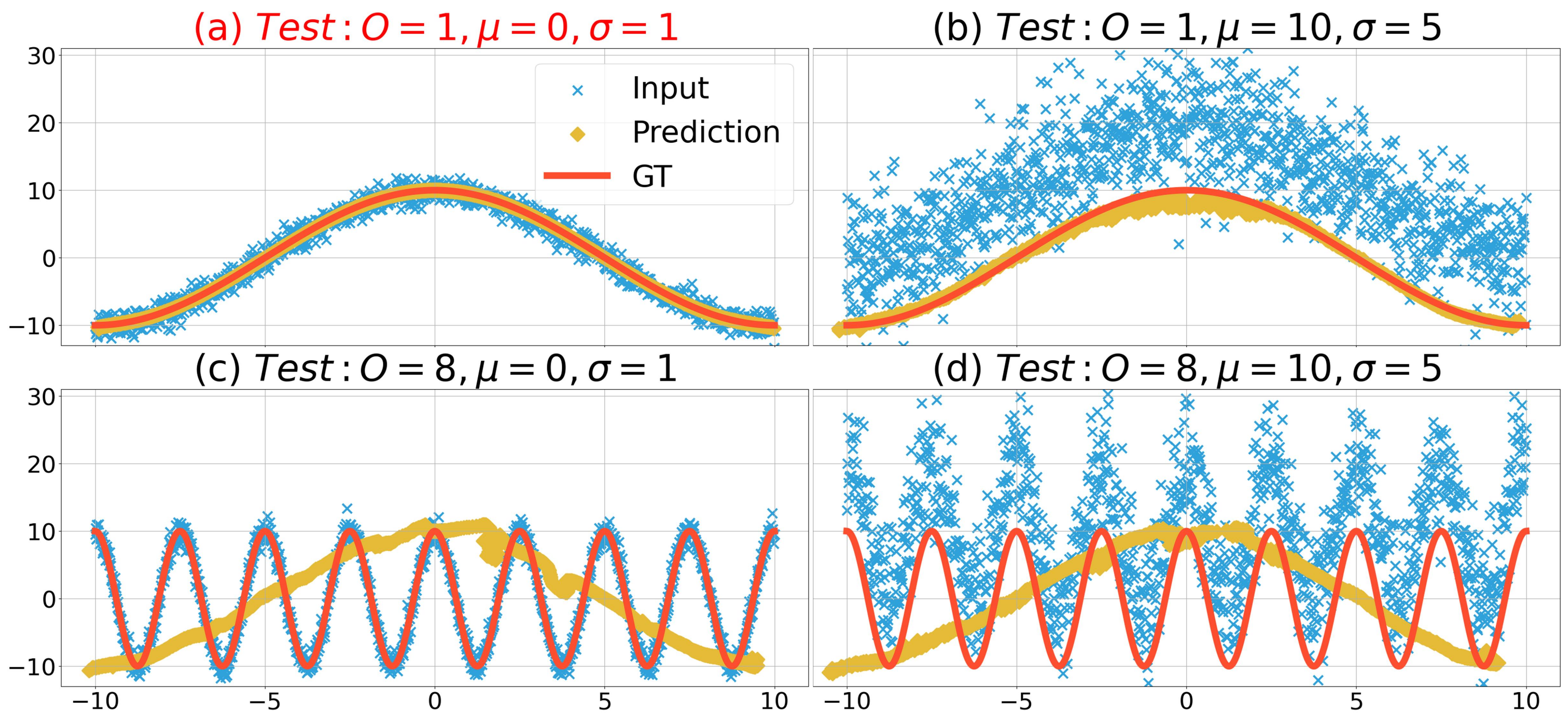}
\vspace{-5mm}
    \caption{Four test results of the network trained with the setting $O=1, \mu=0, \sigma=1$. Red titles indicate that the testing and training settings are the same.}
    \label{fig:demoO1}
\vspace{-3mm}
\end{figure}
The analogy between this function denoising task and image restoration is summarized in Table \ref{tab:toy_example}.
The function's order determines its complexity, corresponding to the background content in an image.
The noise parameters $\mu$ and $ \sigma$ control the degradation pattern, analogous to different degradations in image restoration.
During training, we sample function segments, which is analogous to cropping image patches.
Our training set consists of 10,000 such function segments.
Each segment is a random interval of the function with added noise.
For example, a training function segment can be represented as $\{(x, y+n) \mid y = f(x), \ x \in (a, b), n \sim \mathcal{N}(\mu, \sigma)\}.$
Each segment contains 128 sample points, analogous to a fixed-size image patch.
During testing, the entire noisy function is fed into the network \( D \) for denoising, similar to processing a full low-quality image.
We use a small-scale network due to the task's simplicity.
It is a five-layer CNN with roughly 400k parameters and Leaky ReLU activations.
The training loss function is defined as $L = \| (x, y) - D(x, y+n) \|$.

\begin{figure}[t]
    \centering
    \includegraphics[width=0.7\linewidth]{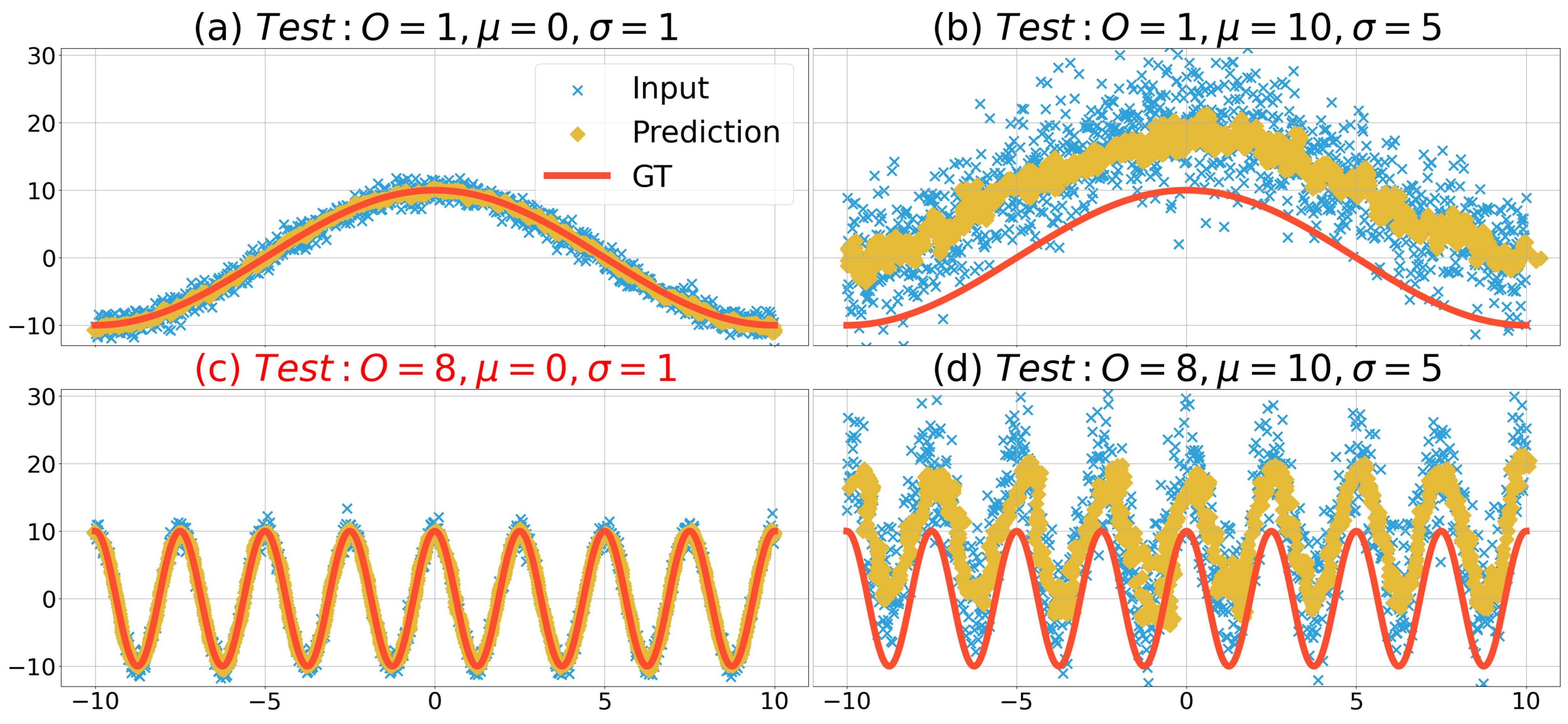}
\vspace{-5mm}
    \caption{Four test results of the network trained with the setting $O=8, \mu=0, \sigma=1$. Red titles indicate that the testing and training settings are the same.}
    \label{fig:demoO8}
\vspace{-3mm}
\end{figure}

\subsection{Fitting the Background Function}

We train the network using a background function with \(O = 1\) and Gaussian noise characterized by \(\mu = 0\) and \(\sigma = 1\).
The GT function becomes $ y = 10\cos(\frac{\pi}{10}x)$ and the noise is sampled from the distribution $\mathcal{N}(0, 1)$.
Fig. \ref{fig:demoO1} shows four test results with different settings. By comparing these results, we can draw the following observations:

    (1) By examining the situation where \textit{the function type remains unchanged while the noise varies} (Fig. \ref{fig:demoO1} (a) and (b)), we observe that the predicted results closely match the ground truth (GT) function. This indicates that the network demonstrates strong generalization ability and effectively removes unknown noise.
    
    (2) When comparing the case where \textit{the function type changes while the noise remains constant} (Fig. \ref{fig:demoO1} (a) and (c)), the network fails to recover the GT cosine function with \(O=8\). Instead, it generates a result similar to the training cosine function with \(O=1\), suggesting an overfit to the training background function.
    
    (3) From the scenario where \textit{both the function type and noise change} (Fig. \ref{fig:demoO1} (a) and (d)), we observe that the network's predictions still resemble the cosine function with \(O=1\). It shows no resemblance to the pattern of the input data points. This further suggests that the training cosine function with \(O=1\) has a significant impact on the network's behavior.

From the above observations, it can be observed that the network has overfitted to the training background function with \(O=1\), while the noise pattern has little influence on the network's predictions.

\begin{figure*}[t]
    \centering
    \includegraphics[width=0.9\linewidth]{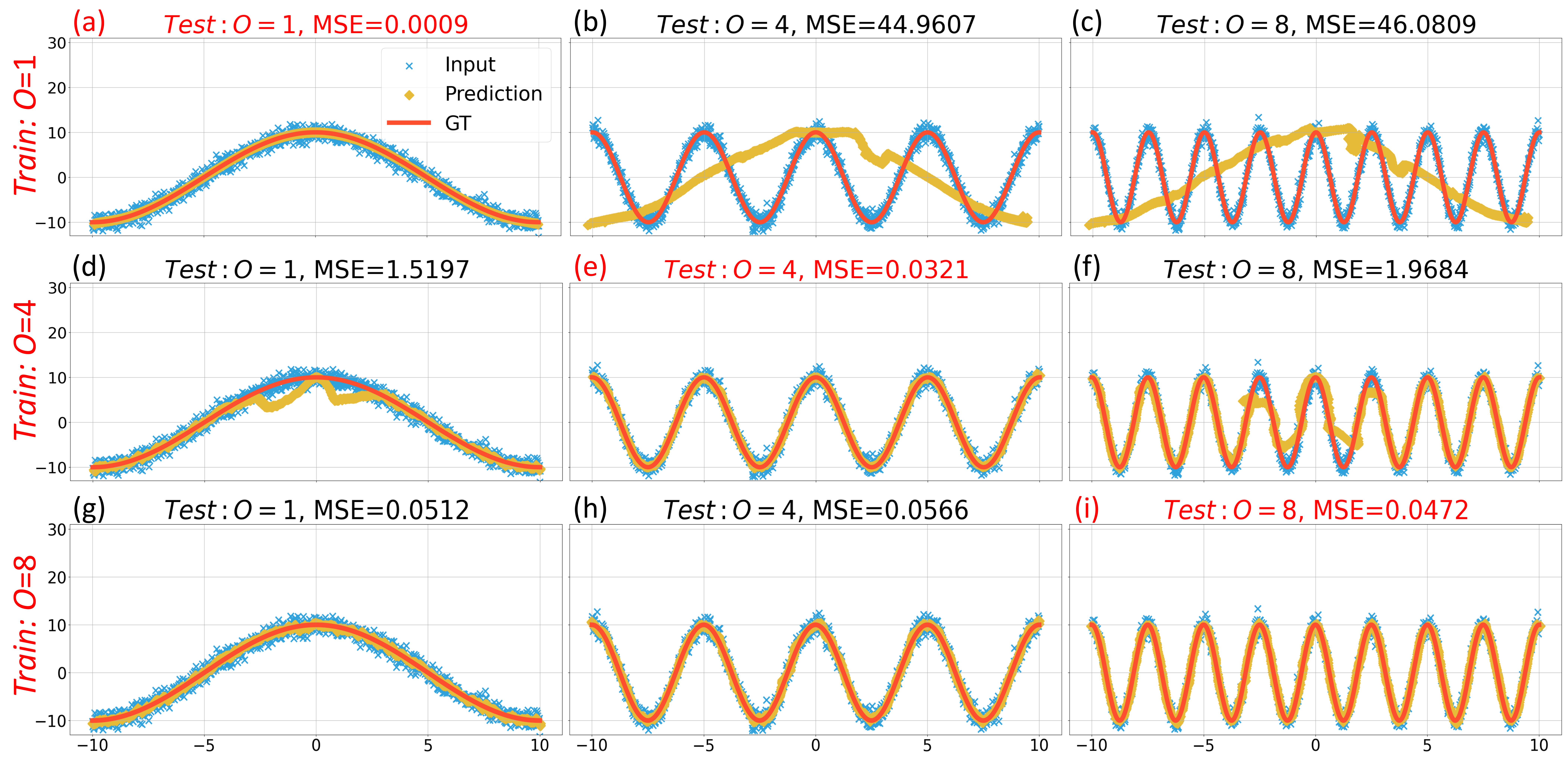}
\vspace{-5mm}
    \caption{Each row, from top to bottom, shows denoising networks trained with cosine function order $O = 1, 4, 8$, where Gaussian noise has $\mu = 0, \sigma = 1$. Each subplot title represents the order used during testing, with Gaussian noise remaining the same, $\mu = 0, \sigma = 1$. The testing Mean Square Error (MSE) is also reported in the title. Red titles indicate that the testing and training settings are the same.}
    \label{fig:all_order}
\end{figure*}

We train separate networks on cosine functions of order \(O = 1, 4, 8\) with fixed Gaussian noise \(\mu = 0\) and \(\sigma = 1\) and evaluate them on various test settings.
The results are shown in Fig.~\ref{fig:all_order}.
Comparing the matched test cases (Fig.\ref{fig:all_order} (a), (e), and (i)), the MSE increases from 0.0009 to 0.0472 as the order increases.
This suggests that higher-order functions are more difficult for the network to learn.
The results reveal that the network's behavior fundamentally changes with the training function's order.
The network trained on \(O=1\) (Figs.\ref{fig:all_order} (a)-(c)) rigidly adheres to the shape of its training function, ignoring the test data's underlying signal.
In contrast, the network trained on the complex $O=8$ function (Figs.\ref{fig:all_order} (g)-(i)) largely ignores its training and instead fits the new test functions.
The $O=4$ model (Figs.\ref{fig:all_order} (d)-(f)) represents a middle ground, with predictions influenced by both the training and test functions.
This creates noticeable fluctuations, as seen in Fig.\ref{fig:all_order} (d), where the output oscillates between the two patterns.
Overall, moving from the top to the bottom row of Fig.~\ref{fig:all_order} reveals a clear behavioral shift.
 \textit{As the order \(O\) of the training background function increases, the network shifts from overfitting to not fitting the training function during inference.}

\begin{figure*}[t]
    \centering
    \includegraphics[width=0.9\linewidth]{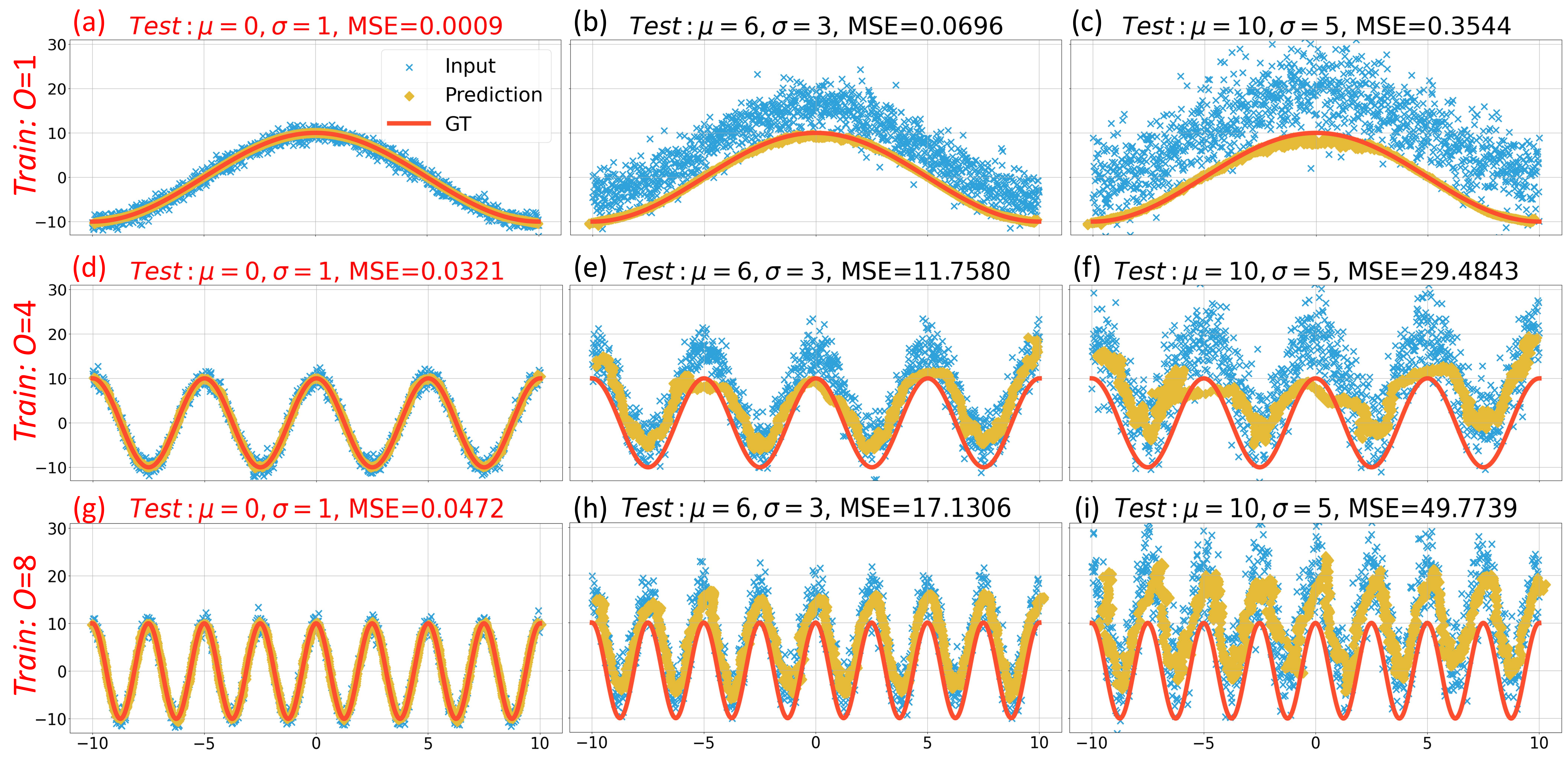}
\vspace{-5mm}
    \caption{Each row from top to bottom shows denoising networks trained with cosine function order $O = 1, 4, 8$, where Gaussian noise has $\mu = 0, \sigma = 1$. Each subplot title represents the Gaussian noise used during testing, with the cosine function order remaining the same as the training one. The testing Mean Square Error (MSE) is also reported in the title. Red titles indicate that the testing and training settings are the same.}
    \label{fig:all_noise}
    \vspace{-3mm}
\end{figure*}

\subsection{Fitting the Additive Noise}
Here, we consider the network training settings with a function order of \(O = 8\) and Gaussian noise characterized by \(\mu = 0\) and \(\sigma = 1\). 
Four results are illustrated in Fig.~\ref{fig:demoO8}. 
We can draw the following observations:

    (1) By comparing the situation where \textit{the function type remains unchanged while the noise varies} (Fig. \ref{fig:demoO8} (c) and (d)), we can observe that the network's prediction closely follows the input noisy data. It does not exhibit the same behavior as the network trained with \(O=1\).
    
    (2) When comparing the case where \textit{the function type changes while the noise remains the same} (Fig. \ref{fig:demoO8} (c) and (a)), we can observe that the network can reconstruct the corresponding GT function. This demonstrates the network's ability to remove Gaussian noise with \(\mu = 0\) and \(\sigma = 1\) across different background functions.
    
    (3) From the scenario where \textit{both the function type and noise change} (Fig. \ref{fig:demoO8} (c) and (b)), the predicted result shows no resemblance to the training function's shape and falls within the noisy data. They deviate from the GT values by approximately 10, which corresponds to the mean of the noise. This indicates that the network behavior is more influenced by the noise.

From the above observations, we can conclude that the network trained with \(O=8\) has overfitted to the training noise.
Its prediction falls within the noise distribution, treating all noise as if it were sampled from $\mathcal{N}(0, 1)$. 
Based on the experiments shown in Fig. \ref{fig:demoO1} and \ref{fig:demoO8}, we observe the following ranking of the network's tendency to overfit during training: first-order cosine function ($O=1$) > Gaussian noise $\mathcal{N}(0, 1)$ > higher-order cosine function ($O=8$).

We now test the models trained on function orders \(O=1, 4, 8\) against various unseen noise patterns, with results shown in Fig.\ref{fig:all_noise}.
The network trained on the simple \( O = 1 \) function generalizes robustly, accurately recovering the GT signal even under heavy, unseen noise (Fig.\ref{fig:all_noise} (a)–(c)).
In contrast, the network trained on the complex \( O = 8 \) function fails to generalize; its output deviates from the GT and follows the noisy input instead (Fig.\ref{fig:all_noise} (g)–(i)).
The  \( O = 4 \) model exhibits an intermediate behavior between these two extremes (Fig.\ref{fig:all_noise} (d)–(f)).
This trend is starkly illustrated by the results under heavy test noise sampled from \( \mathcal{N}(10, 5) \), as seen in Fig.~\ref{fig:all_noise} (c), (f), and (i).
Here, the \( O = 1 \) model maintains excellent performance with an MSE of just 0.3544, while the \( O = 8 \) model fails completely, yielding an MSE of 49.7739.
In summary, \textit{as the order \( O \) of the training background function increases, the network progressively fits the noise pattern instead of the GT background function.}

\subsection{Conclusion of the Analogous Experiment}
In this section, we used an intuitive toy task to analyze the network's generalization behavior.
Our key finding is that the network's strategy is governed by the relative complexity of the background function versus the noise.
In our baseline case—a simple function (\(O=1\)) with simple noise (\(\mu=0, \sigma=1\))—the network overfits the background function.
The test results closely resemble the training function when faced with different noisy data. 
However, as we hold the noise constant and increase the function's complexity, the network's behavior shifts.
It gradually moves from overfitting the background function to overfitting the training noise pattern instead.
Once this shift occurs, the network loses its ability to generalize, failing to remove any noise that deviates from its training distribution.
Based on our observations, \textbf{we conclude that a network exhibits robust generalization when it fits the background content instead of the degradation}.
This allows the network to effectively handle unseen degradation.

\section{Implication}\label{sec:implict}
This paper concludes that the key to improving the generalization ability of low-level vision networks is to guide the network to learn the desired image content.
In this section, we propose some feasible solutions based on this insight.

\subsection{Adjusting Training Set to Avoid Degradation Overfitting} \label{implication1}
Our previous experiments have yielded three significant practical findings:
(1) By limiting the number of background images used in training, the network can focus more on learning the image content instead of overfitting to the rain streaks;
(2) Enlarging the range of rain streaks in the training set can allow for the use of more background images in training;
(3) Surprisingly, a small number of background images is sufficient for reconstruction performance.
These insights can be directly applied to enhance the generalization of existing models with minimal modification.
Our strategy is therefore straightforward: balance the complexity of the backgrounds against the rain to prevent the network from overfitting to the degradation.

\begin{table}[t]
 \caption{Quantitative comparisons. $\uparrow$ means the higher the better while $\downarrow$ means the lower the better.}\label{tab:app}
    \centering
    \huge
    \resizebox{0.8\textwidth}{!}{
        \begin{tabular}{cc|ccc|ccc|ccc}
        \toprule
        \rowcolor[gray]{.9} \multicolumn{2}{c}{Training Objective} & \multicolumn{3}{c}{ResNet} & \multicolumn{3}{c}{SPDNet~\cite{yi2021structure}} & \multicolumn{3}{c}{RCDNet~\cite{wang2020model}} \\
        \rowcolor[gray]{.9} Back. & Range & $E_R$ $\uparrow$ & $E_B$ $\downarrow$ & PSNR $\uparrow$ & $E_R$ $\uparrow$ & $E_B$ $\downarrow$ & PSNR $\uparrow$ & $E_R$ $\uparrow$ & $E_B$ $\downarrow$ & PSNR $\uparrow$ \\
        \midrule
        30k & Medium & 31.24 & 10.79 & 25.15 & 33.63 & 5.49 & 30.51 & 26.55 & 5.41 & 28.54 \\
        \midrule
        64 & Medium & 53.33 & 25.02 & 20.87 & -- & -- & -- & 45.47 & 14.78 & 25.32 \\
        512 & Large & -- & -- & -- & 39.88 & 8.91 & 28.57 & 37.53 & 7.16 & 29.60\\
        256 & Large & 45.64 & 16.51 & 24.30 & 38.87 & 8.03 & 29.40 & 40.40 & 8.52 & 29.08 \\
        128 & Large & 51.75 & 23.53 & 21.45 & 43.20 & 14.59 & 25.67 & 44.67 & 13.72 & 26.09\\
        \bottomrule
        \end{tabular}
        }
        \hfill
\end{table}

Some quantitative results are presented in Tab. \ref{tab:app}.
We demonstrate our strategy on three baselines: ResNet, SPDNet \cite{yi2021structure}, and RCDNet \cite{wang2020model}.
These models are conventionally trained on 30,000 backgrounds with medium-range rain and tested on the R100 dataset \cite{yang2017deep}.
We evaluate using our decoupled metrics, $E_R$ for rain removal and $E_B$ for background reconstruction, with PSNR as a reference.
It can be seen that using the existing training methods cannot generalize well to the unseen rain of R100, which is shown by the poor deraining performance in Tab. \ref{tab:app}.
However, due to the learning on a large number of images, the reconstruction errors of the baseline models are generally lower.
Thus the PSNR values cannot objectively reflect the rain removal effect.
Applying our first strategy, we reduce the training backgrounds to 64 images.
At this time, the rain removal performance has greatly improved, but at the cost of background reconstruction performance.
By balancing a more complex rain range with more background images, we achieve a strong trade-off, excelling in both rain removal and background reconstruction.

\begin{figure*}[t]
    \centering
    % \vspace{-5pt}
    \includegraphics[width=\linewidth]{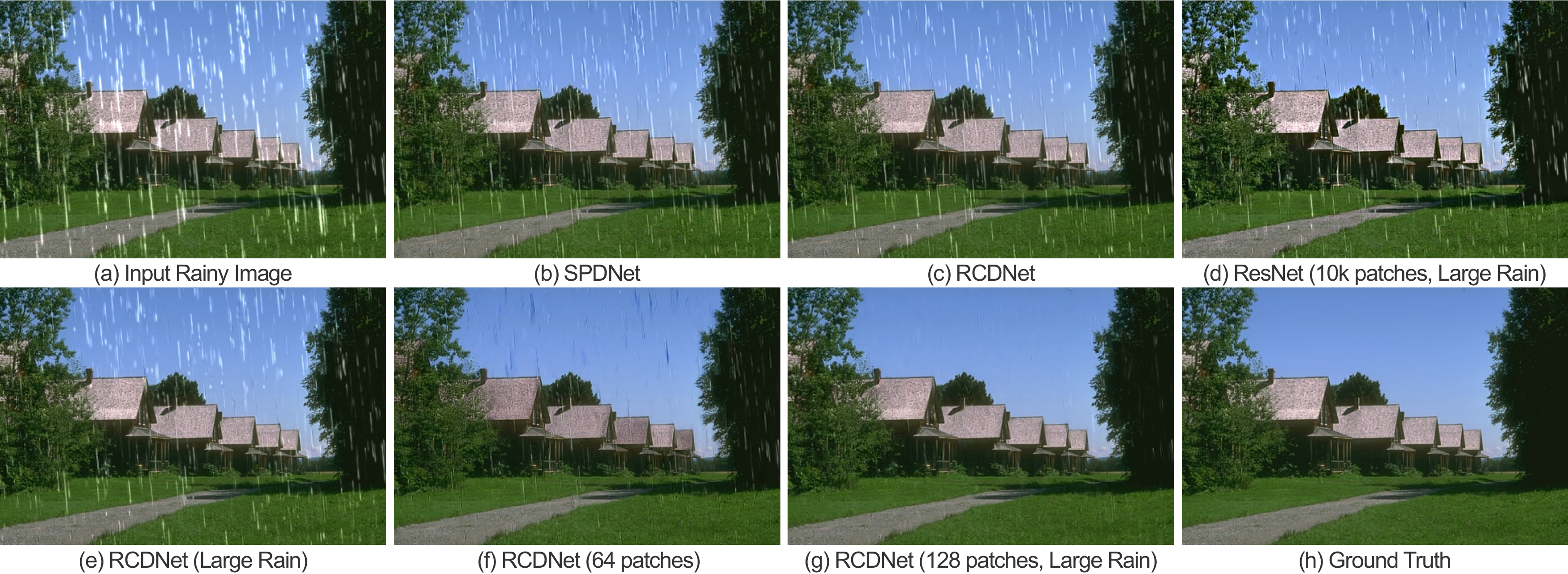}
    \vspace{-12mm}
    \caption{Visualization of the deraining results on a synthetic image. Zoom in for better comparison.}
    \label{fig:compare}
\vspace{-3mm}
\end{figure*}

\begin{figure*}[t]
    \centering
    % \vspace{-5pt}
    \includegraphics[width=\linewidth]{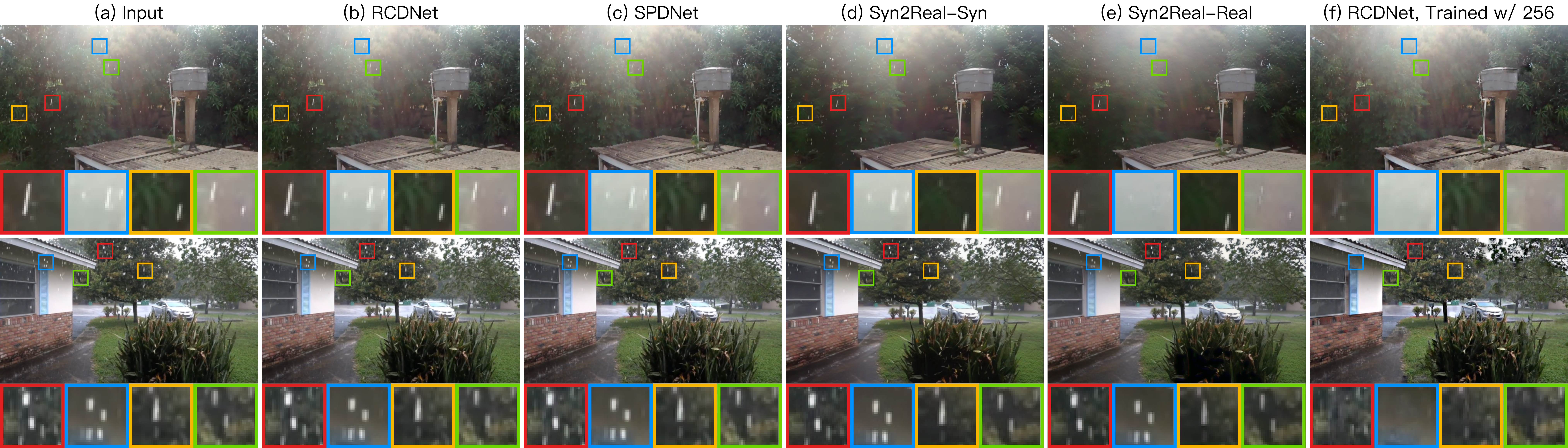}
    \vspace{-12mm}
    \caption{Qualitative results on real-world test images. Zoom in for better comparison.}
    \label{fig:real}
   \vspace{-3mm}
\end{figure*}

\figurename~\ref{fig:compare} shows qualitative comparisons of these models under different training objectives.
First, we observe that the baseline models for SPDNet and RCDNet fail to generalize, despite their advanced architectures.
Using a larger range of rain can bring limited improvements.
Using fewer backgrounds (64 patches) with medium rain significantly improves deraining, but at the cost of unstable image reconstruction.
In contrast, by balancing a larger rain range with 128 background patches, the model excels at both rain removal and background reconstruction.
Crucially, these significant improvements were achieved solely by adjusting the training patch numbers.
No changes were made to the network architectures.

The comparison on real images is shown in \figurename~\ref{fig:real}.
For context, we compare against semi-supervised methods like Syn2Real \cite{yasarla2020syn2real,yasarla2021semi}, which also aim to improve performance on real images.
Syn2Real-Syn is trained on synthetic data, and Syn2Real-Real is trained on synthetic labeled data and real unlabeled data.
As expected, models trained purely on synthetic data fail on real images due to the domain gap in rain patterns.
When obtaining some real images, Syn2Real-Real can indeed achieve some improvement. However, these improvements are not brought by improving the generalization ability.
It works by effectively converting "unseen rain" into "seen rain" by incorporating it into the training process.
This approach is impractical, as collecting even unlabeled real data is extremely difficult.
Our method improves generalization performance and achieves better results on test images.

\begin{figure}[t]
    \centering
    \includegraphics[width=0.7\linewidth]{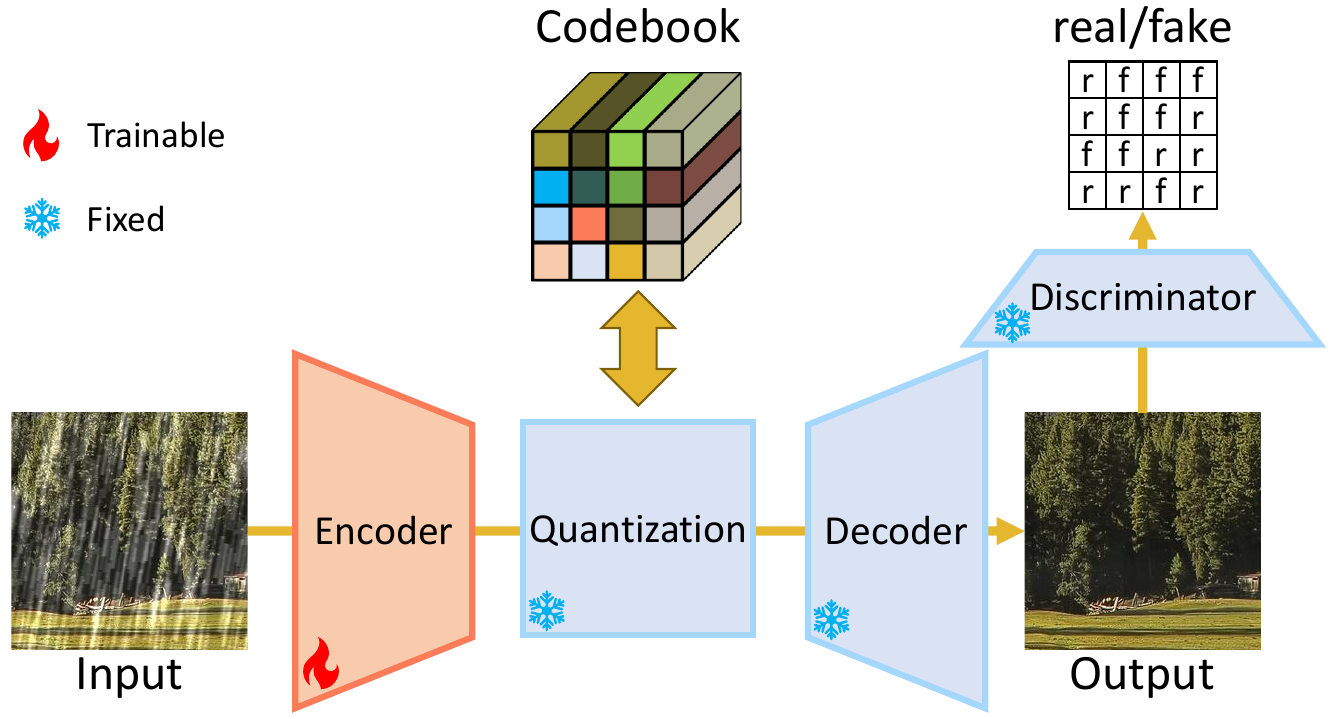}
    \vspace{-5mm}
    \caption{Fine-tuning strategy of image draining with pre-trained VQGAN. }
    \label{fig:finetune}
    \vspace{-3mm}
\end{figure}

\subsection{Empowering Networks to Leverage Image Content Priors} \label{sec:prior}
Our key insight is that strong generalization arises when a network learns to fit image content instead of degradation.
In Sec. \ref{implication1}, we achieved this by balancing the learning difficulty between the background and rain to guide a network trained from scratch.
While balancing the complexity of training data is effective, manually controlling the ratio between content and degradation is practically challenging. To address this, we directly leverage strong, pre-trianed priors of natural image content.
This forces the network to utilize the content prior, eliminating the need for the careful balancing.

In recent years, image restoration techniques based on generative models have been emerging rapidly \cite{liu2024adaptbir,lin2024diffbir,supir}. 
Compared to traditional methods, these approaches not only generate more realistic image details but also exhibit significantly better generalization. 
However, the reason behind this improved generalization has not yet been thoroughly discussed in existing work.
Based on the findings of this paper, we suggest that the superior generalization of recent generative model-based image restoration methods can be attributed to their strong image content priors. 
Therefore, to enhance the network's generalization ability, leveraging content priors from pre-trained models is a promising idea. 
We conduct extensive experiments to validate the effectiveness of this strategy.

VQGAN \cite{esser2021taming} is a representative and fundamental generative model. 
It features an encoder-decoder architecture where the encoder compresses the input image into a lower-dimensional latent space, which is then quantized using vector quantization.
The codebook in VQGAN, a finite set of learned vectors, plays a crucial role in the vector quantization process. 
It enables the model to convert continuous latent features into discrete representations by mapping these features to the closest vectors in the codebook. 
Each vector corresponds to a distinct feature or characteristic derived from the image dataset. 
After quantization, the image is represented as a series of discrete codebook vectors that compress the original image content.
In this sense, the codebook can be regarded as an abstract content prior, capturing natural image features.

\begin{figure*}[t]
    \centering
    \includegraphics[width=\linewidth]{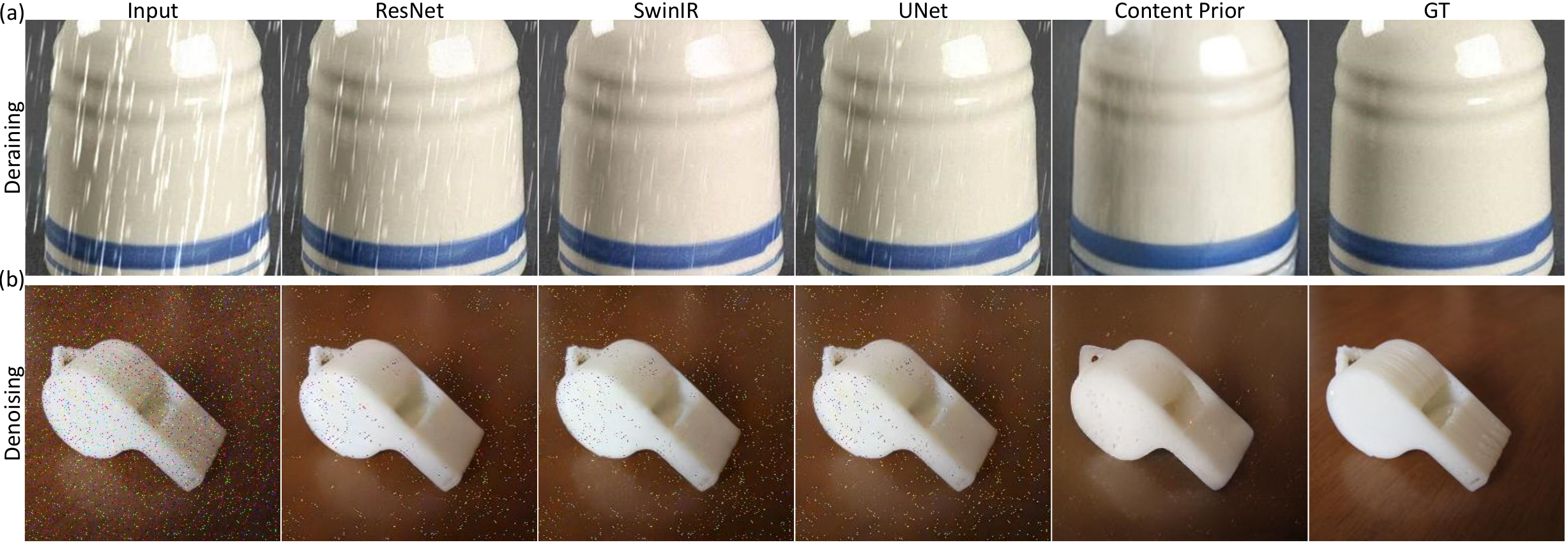}
    \vspace{-10mm}
    \caption{Visual comparisons across ResNet, SwinIR, UNet, and the fine-tuned VQGAN with content prior for image deraining and denoising tasks. }
    \label{fig:VQ_compare}
    \vspace{-3mm}
\end{figure*}

\begin{figure*}[t]
    \centering
    \includegraphics[width=\linewidth]{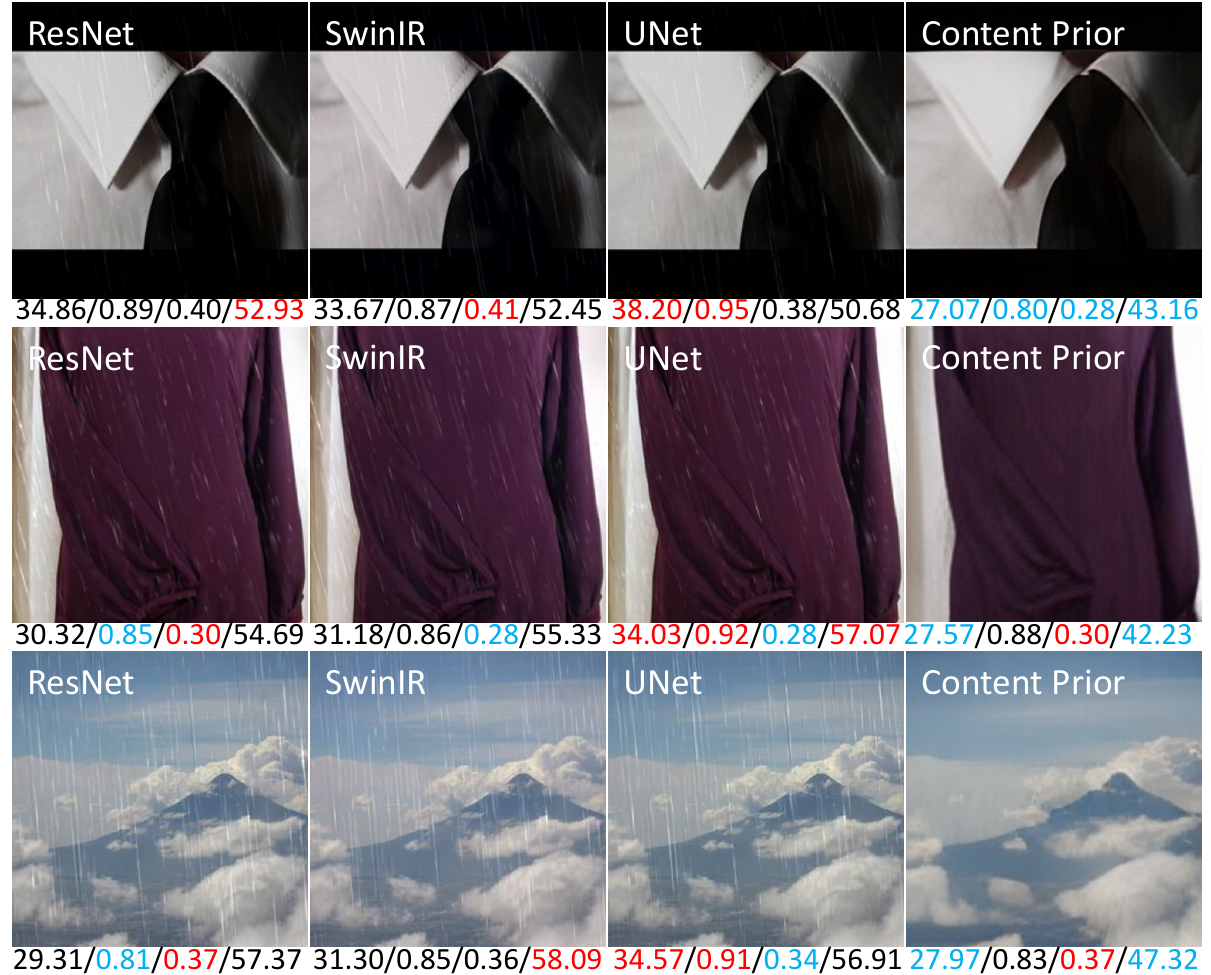}
    \vspace{-10mm}
    \caption{\mod{Quantitative comparison using PSNR/SSIM/MANIQA/MUSIQ. Optimal values are in red, while the poorest are in blue. The discrepancy highlights that traditional IQA scores do not objectively reflect the generalization capability of low-level vision models.}}
    \label{fig:iqa_failure}
    \vspace{-3mm}
\end{figure*}

\begin{figure*}[t]
    \centering
    \includegraphics[width=0.7\linewidth]{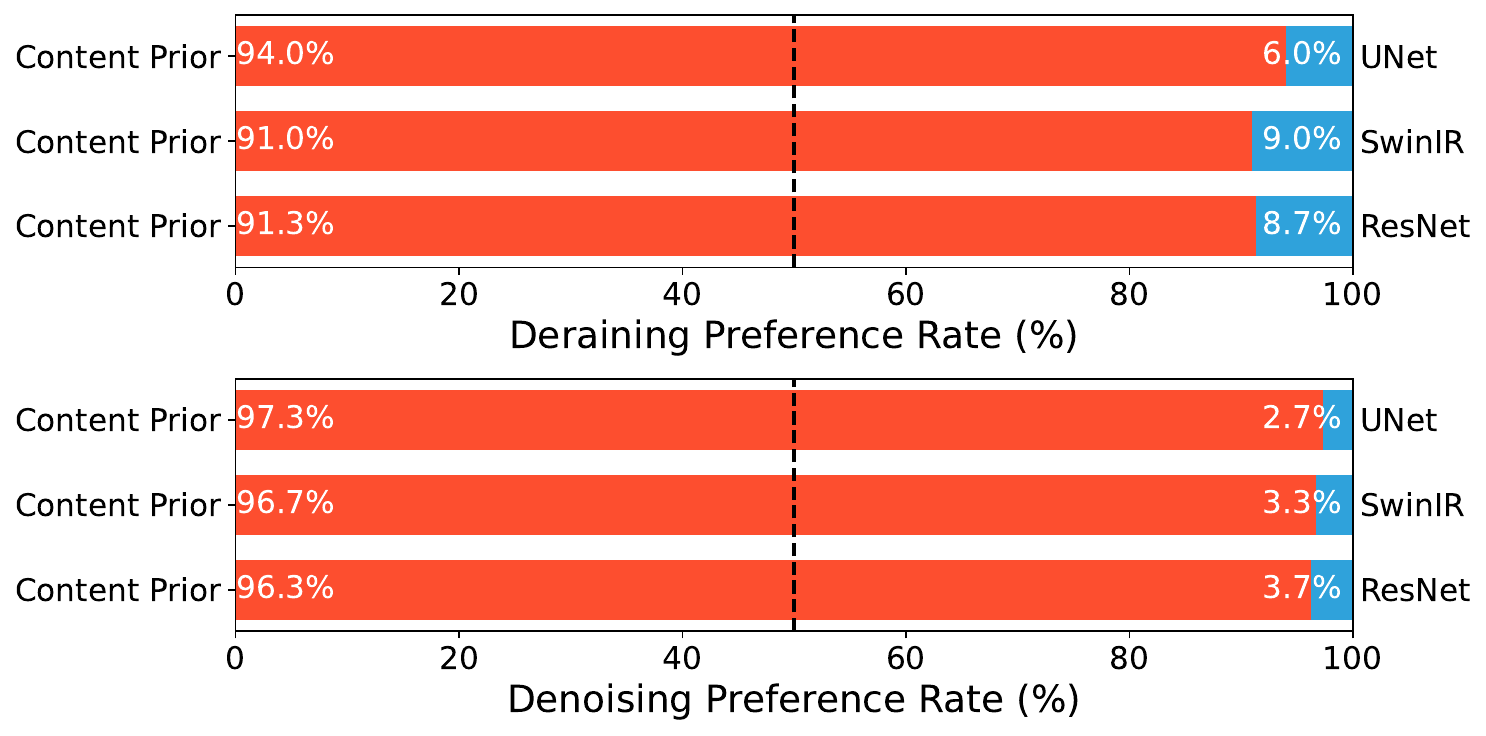}
    \vspace{-5mm}
    \caption{The preference results indicates that the content prior method outperforms others in both deraining and denoising tasks.}
    \label{fig:DepictQA}
    \vspace{-3mm}
\end{figure*}

\begin{figure*}[t]
    \centering
    \includegraphics[width=\linewidth]{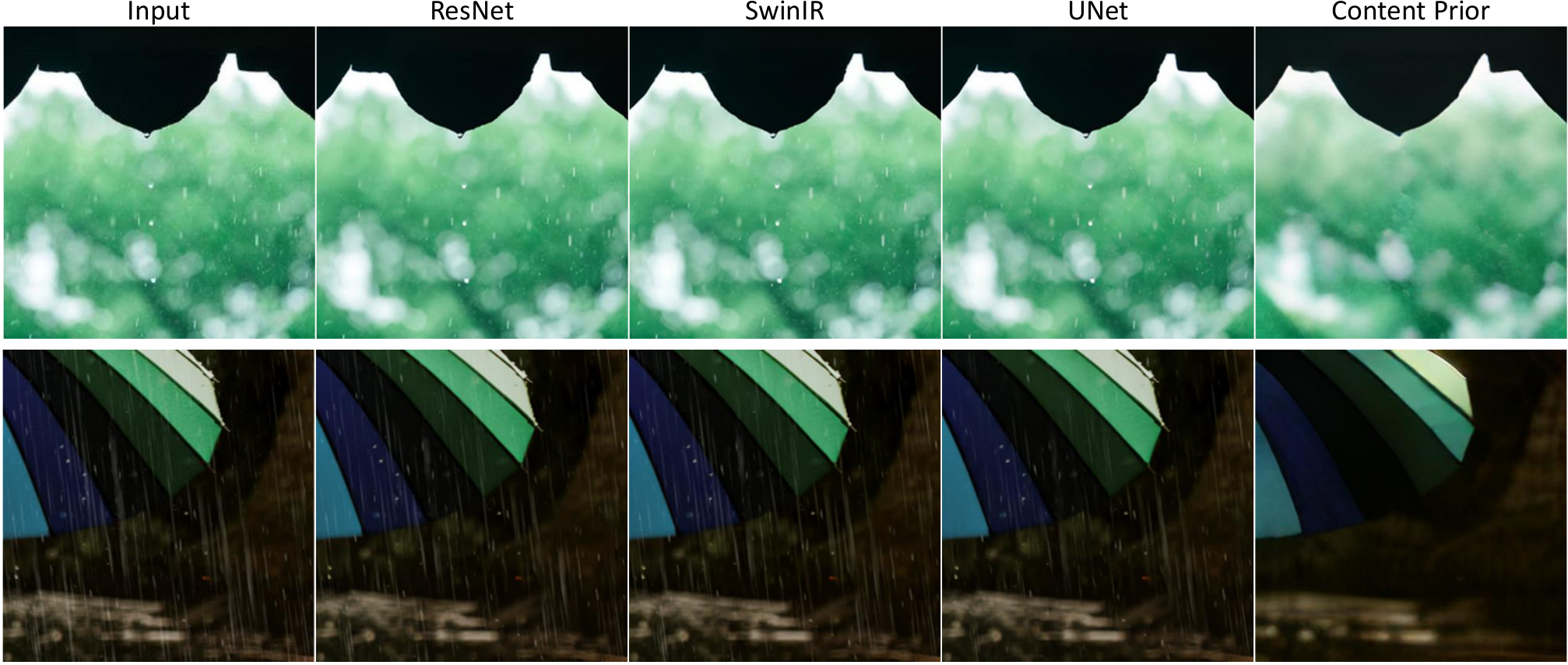}
    \vspace{-10mm}
    \caption{Visual comparisons across ResNet, SwinIR, UNet, and the fine-tuned VQGAN with content prior in real-world image deraining.}
    \label{fig:Real_VQ}
    \vspace{-3mm}
\end{figure*}

\mod{We leverage a VQGAN pre-trained on ImageNet to utilize its robust natural image manifold. By fine-tuning only the encoder while freezing the codebook and decoder modules (\figurename~\ref{fig:finetune}), the network is forced to map degraded inputs into a pre-learned clean latent space. For implementation, we used 10,000 ImageNet image pairs for 50 epochs, trained on two NVIDIA RTX A6000 GPUs with a batch size of 14. In contrast, baselines (ResNet, SwinIR, and UNet) were trained from scratch on 30,000 images for 100,000 iterations. 
As shown in \figurename~\ref{fig:VQ_compare}(a), traditional architectures struggle with unfamiliar Rain100L patterns, leaving visible artifacts. Conversely, the content-prior approach demonstrates superior generalization; by prioritizing structural manifolds over degradation patterns, it effectively removes rain streaks even when they were not encountered during the training stage.}

Furthermore, we extend this strategy to the image denoising task for broader validation. In this experiment, ResNet, SwinIR, and UNet are trained for 100,000 iterations using 30,000 images corrupted by Gaussian noise $\mathcal{N}(0, 30)$, while the VQGAN is fine-tuned using a subset of 10,000 images for 50 epochs. To evaluate generalization, we tested these models on 300 images containing salt-and-pepper noise (2\%), a degradation type unseen during training. As illustrated in \figurename~\ref{fig:VQ_compare} (b), traditional architectures trained from scratch struggle to eliminate these unfamiliar noise patterns, leaving heavy artifacts. In contrast, the method incorporating content priors effectively reconstructs clean images.

However, quantitatively assessing such generalization performance remains a significant challenge \cite{gu2023networks,gu2020pipal}. As shown in \figurename~\ref{fig:iqa_failure}, traditional full-reference metrics (PSNR, SSIM \cite{ssim}) and even recent no-reference metrics (MANIQA \cite{maniqa}, MUSIQ \cite{musiq}) often produce counter-intuitive results. In many instances, baseline models achieve higher IQA scores simply by perfectly preserving the background while failing to remove any rain streaks. Conversely, the content-prior method, despite effectively removing degradations, may receive lower scores due to minor pixel-level shifts inherent in generative reconstruction. This discrepancy reveals that standard IQA scores cannot objectively reflect true deraining performance or generalization capability.

To circumvent these limitations, we utilize the vision-language model DepictQA \cite{you2023depicting} to evaluate model performance from a perceptual perspective. Specifically, we fine-tune DepictQA to focus exclusively on the effectiveness of artifact removal. As shown in \figurename~\ref{fig:DepictQA}, the content-prior method significantly outperforms other approaches in both deraining and denoising tasks, achieving preference rates exceeding 90\%. These results demonstrate that when evaluated based on the actual removal of unseen degradations, the content-prior approach offers a robust advantage that traditional metrics fail to capture.

The effectiveness of this approach is further validated on real-world image deraining. \figurename~\ref{fig:Real_VQ} illustrates the comparative results on real-world rainy images. Traditional networks trained on synthetic data demonstrate limited effectiveness, often producing outputs nearly indistinguishable from the inputs. In contrast, the content-prior method removes entirely new and complex real rain patterns effectively. This confirms that prioritizing content learning via generative manifolds is the fundamental solution to the generalization bottleneck in low-level vision.

\begin{figure}[t]
    \centering
    \includegraphics[width=1\linewidth]{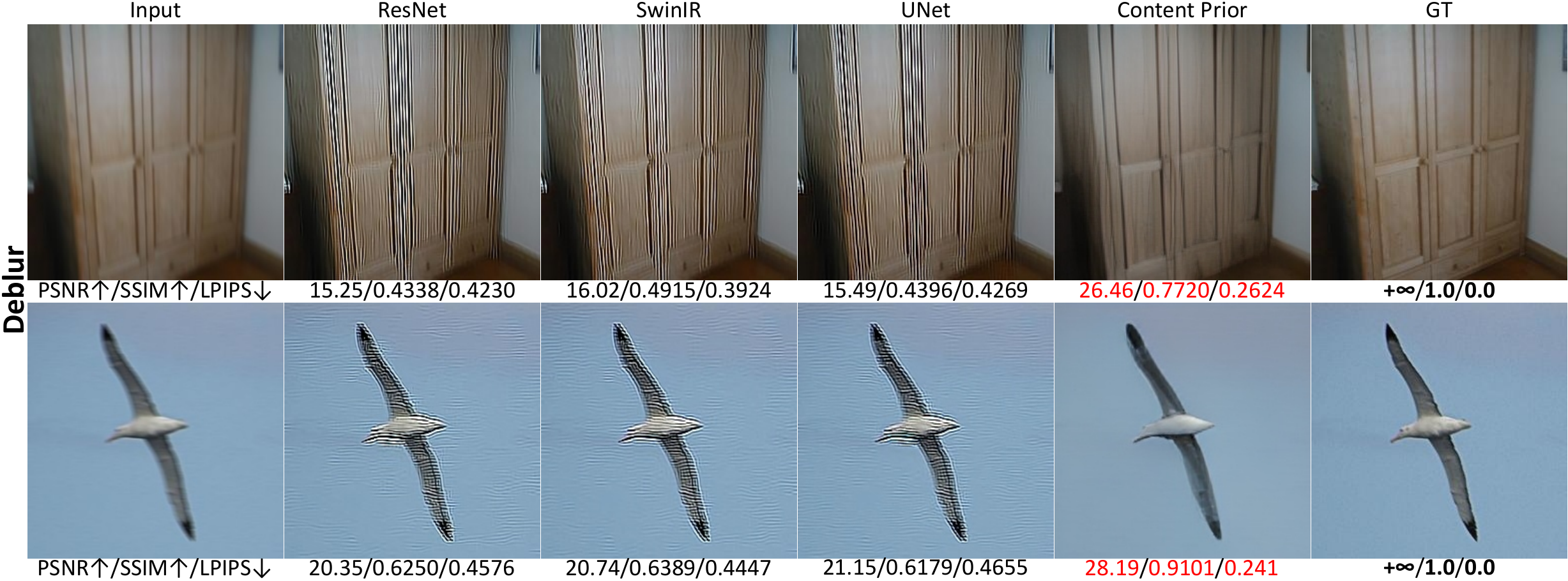}
    \vspace{-10mm}
    \caption{\mod{Visual comparison of image deblurring results under cross-domain scenarios (Training on Gaussian blur vs. Testing on motion blur). Traditional models fail to generalize and produce artifacts.}}
    \label{fig:deblur_visual}
    \vspace{-3mm}
\end{figure}

\mod{To further demonstrate the broader applicability of our findings, we extend our validation to the image deblurring task, moving beyond simple additive models to explore degradations that are inextricably coupled with global image content \cite{he2025diffusion}. Unlike deraining or denoising, deblurring involves a spatial convolution operation where the degradation kernel interacts with the underlying pixels, posing a more challenging test for model generalization. In this experiment, the models are trained on isotropic Gaussian blur with kernel sizes of $\{5, 7, 9\}$, and evaluated on anisotropic motion blur with lengths of $\{7, 9, 11\}$ and random directions. As illustrated in \figurename~\ref{fig:deblur_visual}, traditional architectures (ResNet, SwinIR, and UNet) fail to generalize to these unseen motion patterns, often producing severe ringing artifacts or failing to restore sharp edges. In contrast, the method leveraging content priors effectively reconstructs structures even under out-of-distribution blur kernels.}

\begin{figure}[t]
    \centering
    \includegraphics[width=1\linewidth]{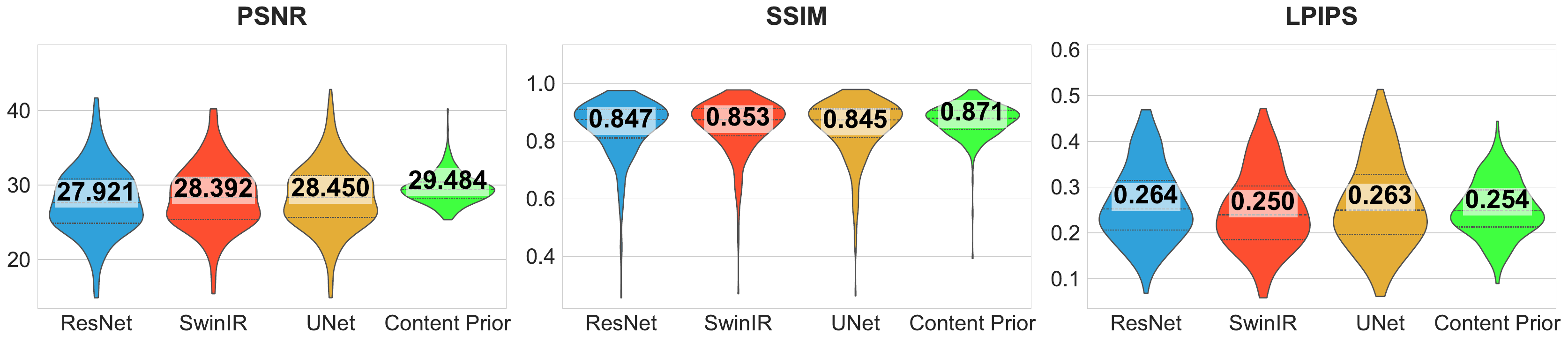}
    \vspace{-10mm}
    \caption{\mod{Violin plots of PSNR, SSIM, and LPIPS for the deblurring task. The numbers represent the mean values. Our method consistently shifts the performance distribution manifold toward the better end and exhibits superior statistical stability compared to baselines.}}
    \label{fig:deblur_metric}
    \vspace{-3mm}
\end{figure}

\mod{Quantitatively, deblurring differs from additive tasks like deraining because it is a global operation; thus, traditional IQA metrics can more effectively reflect the success of generalization and the restoration of structural details. To provide a rigorous statistical assessment, we present the violin plots for PSNR, SSIM, and LPIPS \cite{lpips} in \figurename~\ref{fig:deblur_metric}. Our content-prior method achieves a substantial margin in PSNR (reaching 29.484 dB, roughly 1.0 dB higher than baselines) and the highest SSIM (0.871). More importantly, the shape of the violins reveals that while baseline models exhibit long tails reaching very low performance levels, indicating frequent failures on complex image manifolds, our method maintains a significantly higher performance lower-bound and a more compact distribution. Regarding LPIPS, although SwinIR shows a competitive average, its distribution is highly dispersed, whereas our method provides a consistent generalization anchor. These statistical results empirically validate that forcing the network to prioritize content learning is a robust strategy for handling coupled, global degradations.}

\begin{figure}[t]
% \vspace{-45pt}
    \centering
    \includegraphics[width=1\linewidth]{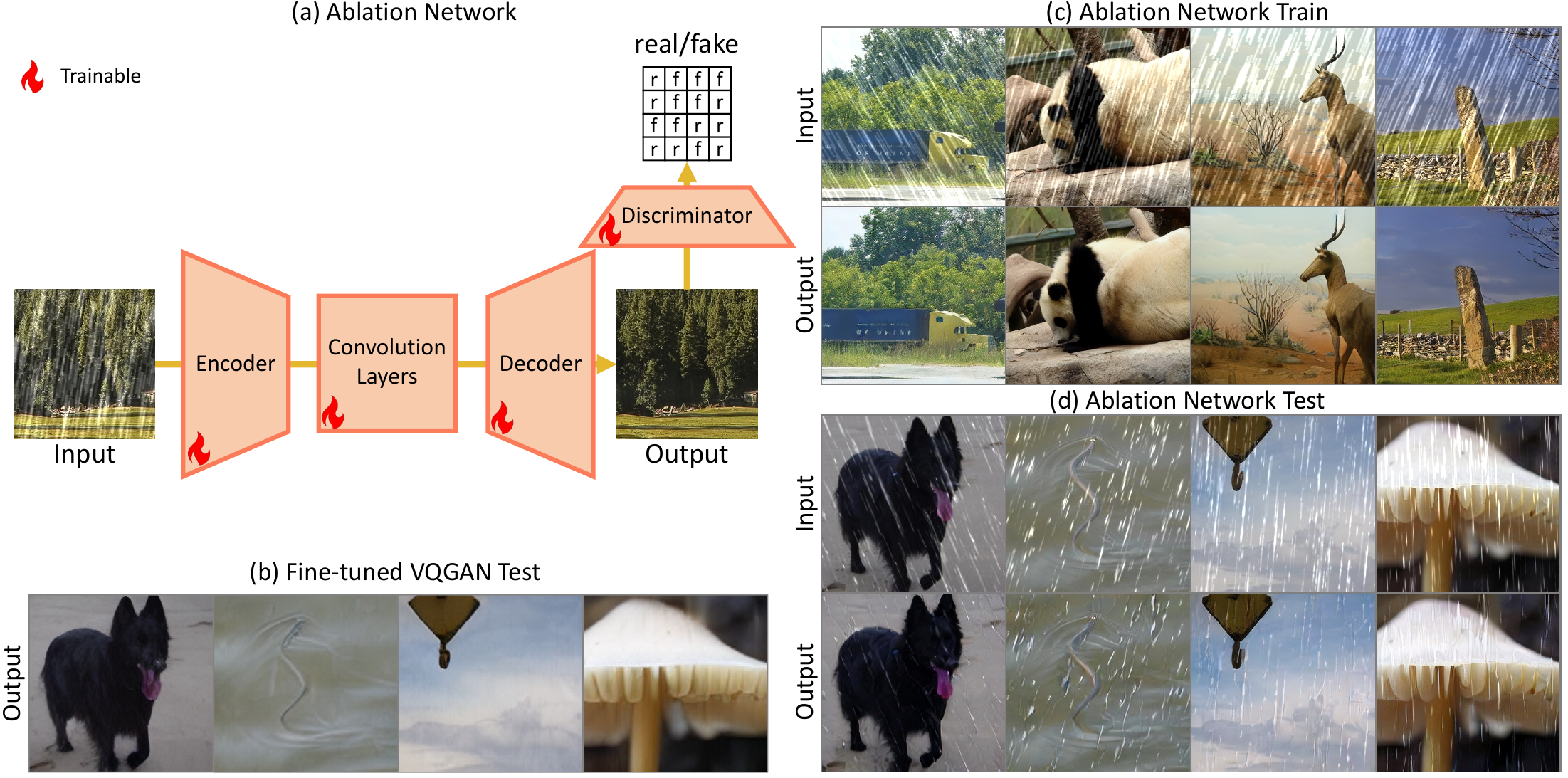}
    \vspace{-10mm}
    \caption{\mod{\textbf{(a)} Ablation study of training from scratch without vector quantization.
\textbf{(b)} The generalization performance of the fine-tuned VQGAN, as depicted in Fig.~\ref{fig:finetune}.  \textbf{(c)} Training performance of a VQGAN without codebook learning.\textbf{(d)} The generalization performance of the ablation network.} }
    \label{fig:Ablation_generalization}
\vspace{-3mm}
\end{figure}

During the fine-tuning stage, we trained only the encoder while keeping other modules fixed to develop a VQGAN-based image restoration approach, as illustrated in \figurename~\ref{fig:finetune}.  
This strategy compels the network to rely on the pre-trained codebook, which encodes features of the natural image space.  
We argue that this approach allows the network to better utilize image content features, leading to enhanced generalization performance.
To further verify this, we design a simple ablation network to observe the impact of incorporating content priors on the network's generalization performance.
Specifically, the general VQGAN network structure is adopted, but the vector quantization strategy is omitted. 
Thus, the ablation network is trained from scratch as a comparative experiment, as illustrated in \figurename~\ref{fig:Ablation_generalization} (a).
The training and testing image deraining results are shown in \figurename~\ref{fig:Ablation_generalization} (c) and (d). 
It can be seen that the ablation network performs very well on the training dataset, effectively removing all rain streaks.
However, when faced with unseen test rain streaks, the VQGAN architecture exhibits behavior similar to traditional networks, failing to remove unfamiliar rain streaks. 
The input rain streaks no longer retain their original shape but become part of the image content.
In the contrast, fine-tuned VQGAN can still handle the unseen testing rain patterns as shown in Fig.\ref{fig:Ablation_generalization} (b).
This ablation study further demonstrates that the generalization improvement indeed stems from the latent image content and features within the VQGAN's codebook.
The credit does not belong to the generative adversarial network training pipeline or the stacking of various modules within the architecture.

\section{Limitation and Conclusion}\label{sec:conclusion}

\mod{\textbf{Limitation:}
Despite superior generalization, the content-prior method presents several limitations. Compared to lightweight end-to-end networks, it incurs higher computational overhead and memory requirements, increasing inference latency. Furthermore, its efficacy depends on the pre-trained prior's quality and domain coverage, the acquisition of which demands significant data and computing resources, posing challenges for resource-constrained deployment. Additionally, generative reconstruction may introduce minor pixel-level shifts or spatial misalignments. Although visually imperceptible, these deviations penalize traditional fidelity metrics like PSNR and SSIM, creating a gap between quantitative evaluation and perceived quality.

\textbf{Conclusion:} This paper investigates the generalization bottleneck in low-level vision through a systematic analysis of image deraining and denoising. We reveal that generalization failure is driven by a ``complexity competition'' mechanism, where networks adopt shortcut learning by overfitting simpler degradation patterns instead of capturing complex image manifolds.

The primary strength of our work is the interpretability-driven perspective that challenges the conventional reliance on blindly scaling up training data. We provide a principled framework for complexity balancing and demonstrate that leveraging generative content priors can effectively break the generalization bottleneck. While generative priors offer superior generalization, they involve higher computational costs and can introduce minor spatial shifts that penalize traditional metrics like PSNR.

These insights benefit the community by providing clear guidelines for more effective dataset curation and encouraging a shift toward content-centric model designs. Future research should focus on developing automated algorithms to optimize the complexity balance and exploring more powerful generative foundations, such as diffusion-based priors. Furthermore, establishing specialized evaluation metrics for generalization remains a critical priority for the field. We hope our findings inspire more principled approaches to building robust and reliable low-level vision systems.}

\section{Acknowledgments}
This work was supported in part by the National Natural Science Foundation of China (Grant No. 62276251), and the Joint Lab of CAS-HK and RGC Early Career Scheme (ECS) No. 24209224.

 \bibliographystyle{elsarticle-num} 
 \bibliography{ref}

\end{document}